# Toward Fully Autonomous Flexible Chunk-Based Aerial Additive Manufacturing: Insights from Experimental Validation

Marios-Nektarios Stamatopoulos*, Jakub Haluska, Elias Small, Jude Marroush, Avijit Banerjee, George Nikolakopoulos

*Robotics & AI Team, Department of Computer, Electrical and Space Engineering, Luleå University of Technology, Luleå SE-97187, Sweden*
*Corresponding author email: marsta@ltu.se*

**Abstract**

A novel autonomous chunk-based aerial additive manufacturing framework is presented, supported with experimental demonstration advancing aerial 3D printing. An optimization-based decomposition algorithm transforms structures into sub-components, or chunks, treated as individual tasks coordinated via a dependency graph, ensuring sequential assignment to UAVs considering inter-dependencies and printability constraints for seamless execution. A specially designed hexacopter equipped with a pressurized canister for lightweight expandable foam extrusion is utilized to deposit the material in a controlled manner. To further enhance precise execution of the printing, an offset-free Model Predictive Control mechanism is considered compensating reactively for disturbances and ground effect during execution. Additionally, an interlocking mechanism is introduced in the chunking process to enhance structural cohesion and improve layer adhesion. Extensive experiments demonstrate the framework's effectiveness in constructing precise structures of various shapes, while seamlessly adapting to practical challenges, proving its potential for a transformative leap in aerial robotic capability for autonomous construction. A video with the overall demonstration can be found here: https://youtu.be/WC1rLMLKEg4.

# 1. Introduction

In recent times, ground breaking advancement in additive manufacturing, seamlessly integrated with autonomous robotics, are unlocking an exciting frontier in next generation construction and manufacturing process. Additive manufacturing has demonstrated a paradigm shift impact, addressing complex manufacturing processes with unprecedented precision and efficiency. Its transformative potential is becoming increasingly evident as it evolves and finds applications across a wide range of industries [1, 2, 3], while simultaneously paving the way for further innovations in the future. An intriguing development is its recent integration into the construction industry, capitalizing on its ability to automate construction processes, provide extensive design flexibility, and construct intricate structures designed using Computer-Aided Design (CAD) software [4, 5]. Numerous studies have demonstrated the design and deployment of large-scale robotic arms and gantry systems for printing building components and even entire houses using a variety of base materials [6]. A key advantage of such methods is their ability to adapt with high level of automation throughout the construction process, making them particularly well-suited for deployment in remote, inaccessible, and harsh environments[7, 8]. Notable examples include disaster-stricken areas, such as regions impacted by fires and earthquakes, where the rapid construction of shelters and basic infrastructure is imperative. The implication of autonomous construction technology is significant for extraterrestrial settlements where immediate human presence is minimal or nonexistent [9]. However, a significant challenge facing both these emerging techniques and conventional construction methods is the requirement for existing infrastructure to facilitate the transportation logistics of the necessary machinery on site and their installation [10]. On top of that, the dimension and the geometry of their available workspace is bounded by their scale, rendering the scalability of the overall structure very difficult and tedious.

Drawing inspiration from nature and biological creatures like bees and wasps in the construction of hives, recent research is exploring the prospect of utilizing the latest innovations in Unmanned Aerial Vehicles (UAVs) and robotics technology, specifically within the field of aerial additive manufacturing, a pioneering concept that is currently in its early conceptual developmental stage is being introduced [11, 12, 13]. In this particular approach, the UAVs act as flying construction workers, where portions of the material used to construct the overall structure. The construction process is executed by depositing material carried by UAVs in consecutive layers to build the structure. This groundbreaking concept of employing the aerial autonomous agents in the construction industry has the potential to transform the construction landscape and further enhance the autonomous large-scale construction. Although still in its infancy, this technology holds great potential for further research and development. However, several challenges must be addressed to ensure a seamless process. Key obstacles include the limited payload capacity of UAVs, their shorter battery life compared to traditional construction vehicles, and the complexities of planning and coordinating autonomous missions. These challenges remain critical barriers to overcome.

In accordance with this objective, this article presents an innovative chunk-based fully autonomous aerial additive manufacturing framework designed to utilize UAVs for autonomous construction. The development highlights the associated challenges and provides a step-by-step intricate details along with insights from experimental evaluation. Given a generic shape in the form of a CAD model mesh to be constructed, the framework offers the flexibility to adapt the process to the characteristics of the available UAVs and the geometry of the mesh, transforming the UAVs into a coordinated aerial construction worker. To validate the approach, a custom-designed hexacopter is used, equipped with a pressurized can carrying expandable foam material, serving as the primary construction material for the entire process. A visual depiction of the experimental setup is presented in Fig. 1.

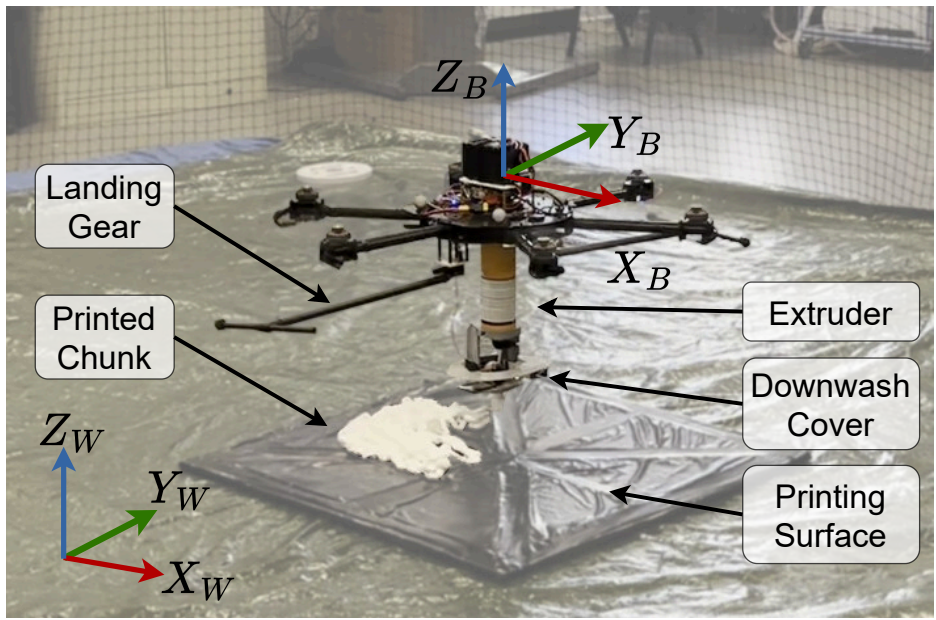

Figure 1: Aerial robotic constructor autonomously building a chunk using the proposed mesh decomposition-based aerial additive manufacturing framework, while extruding material below it.

*1.1. Related Work*

The concept of employing robots to automate the construction process has been a long-standing area of research and remains a significant focus of ongoing studies. The majority of the research so far primarily focused on manipulators supported with ground robots, which either transport and assemble basic building blocks to form a given structure or follow the material deposition based additive manufacturing approach, where material is either carried by the robots or supplied externally [14]. In [15] pioneering research demonstrates multiple ground robots collaboratively constructing structures using stigmergy, an indirect coordination mechanism, by assembling the fundamental stigmergy block. Another study presented in [16] used a manipulator placed on top of a holonomic mobile robot where cementitious material was deposited while simultaneously moving and controlling the arm. This approach enhances flexibility and expands the working envelope. Additionally, aiming to achieve automated printing-in-motion a high-level planning algorithm integrated with a Model Predictive Controller (MPC) is introduced generating robot-base paths in order to avoid collisions and maintain task reachability [17, 18]. Building on a similar concept, [19] presented two mobile ground robots equipped with manipulators that print the same structure concurrently while handling potential collisions between them. Additionally, [20, 21] considered a swarm of ground robots that construct a given structure by splitting it into predefined smaller parts and assigning them to the robots.

In contrast, adoption of aerial robots in autonomous construction is evidently lower compared to the ground-based systems due to their inherent complexity to achieve highly precise motion and associated control challenges. The concept of assembling walls using bricks carried by UAVs was first introduced in [22], where four quadrotors collaboratively assembled a large-scale installation, coordinated and monitored in a centralized approach. Following these initial studies, the concept of autonomous wall building using aerial robots was further investigated by various teams of researchers as part of the globally renowned MBZIRC robotic competition [23, 24, 25]. In another approach towards aerial construction, tensile structures made out of ropes were build by identifying basic building elements of them like knots and ropes and translate them into coordinated quad-copter movements. The robots utilized hybrid force-position control strategies and structures such as bridges were built [26]. The concept of equipping UAVs as workers carrying custom "drone compatible" construction elements is also considered in [27] where four different families of bricks are created and evaluated in a lab. Remote controlled flight were executed and primitive shapes were constructed and tested.

The concept of deploying a flying robot in autonomous construction and additive manufacturing, as a robotic worker carrying extrusion material onboard, is a relatively recent idea, initially presented in a feasibility study [28]. The study highlighted the challenges and limitations of the concept applied in two scenarios, the first for using it as a means to repair damaged structures and the second, to construct new structures. Following this idea, a more intensive study has been presented in [29], which investigated two different types of materials for the performance evaluation. Particularly, expandable polyurethane foam was carried by a single UAV which after making multiple flights it manufactured a large scale tower while a scanner UAV was adjusting the printing process and detecting anomalies and errors in the printing structure. In addition, another UAV equipped with a delta manipulator on its bottom part to reduce positioning inaccuracies deposited a cementitious material in various patterns. The multi-agent scenario was only tested in simulation and in real flight emulations, without physical printing, where printing tasks are assigned in a predefined way UAVs. In our recent work [13, 30, 31], a chunk based framework towards aerial additive manufacturing is presented. In this framework, a shape to be constructed is split into smaller chunks formed using planar cuts, which can be sequentially assigned for printing based on the order of the execution sequences. However, all our previous efforts have been largely constrained to evaluations within simulation or emulation scenarios, primarily focusing on the theoretical validation of the conceptual development, without any real-world implementation of the aerial construction process through the deployment of UAVs in actual 3D printing operations, aimed at physically constructing a structure.

*1.2. Contributions*

In contrast with the existing literature, this article considers a significant leap by introducing the experimental evaluation of a novel mesh-decomposition-based aerial 3D printing framework, making a critical advancement in bridging the gap between simulation and real implementation. The research demonstrates a UAV, precisely maneuvering through designated paths in fully autonomous mode while physically extruding material to manufacture various tangible geometrical shapes. To achieve this milestone, various practical limitations associated with aerial operations are carefully considered, and intricate challenges associated with real-time execution are systematically addressed. Towards this, an aerial hexacopter robotic construction platform is designed and equipped with a pressurized container of lightweight expandable foam material. The extrusion of the printing material is controlled through the onboard computer, enabling precise material deposition during flight. The overall construction pipeline considers an optimal chunking strategy to efficiently decompose the structure into multiple parts, which are systematically organized through a dependency graph to ensure seamless sequential execution. Practical execution challenges, such as ground effect disturbances due to the proximity of the aerial platform to the ground, along with the change in the mass of the overall setup while depleting material during flight, lead to significant deviations of the intended flight trajectory. In order to address this issue, the framework integrates an onboard offset-free Nonlinear Model Predictive Control (NMPC) approach coupled with a reactive Nonlinear Moving Horizon Estimator (NMHE). This control strategy dynamically updates the prediction model in real time, compensating for external disturbances and model mismatches, ensuring precise trajectory fol-lowing throughout the printing process. Another challenge encountered is the slippage of deposited material after extrusion, due to downward wind gusts appearing from the UAV's propellers. While the material's adhesiveness is ideally expected to ensure bonding with the previously deposited layer, in practice, the downwash airflow often disrupts the expected behavior. To address this, an interlocking mechanism is introduced in the chunking process to enhance structural cohesion and improve layer adhesion. Additionally, a protective umbrella-like structure is integrated into the mechanical design to shield the deposition area from propeller-induced wind disturbances. The article provides detailed insights from experimental results, showcasing the framework's successful transition from conceptual development to practical application. The efficacy of the framework is evaluated on various geometric shapes and is critically analyzed on various aspects such as deviations between the desired and the printed chunks, margins of execution errors and influence of physical phenomena such as downwash generated by the propellers while flying and the influence of ground-effects. It concludes with a discussion of future research directions, paving the way for further advancements in aerial additive manufacturing.

## 2. Organization of the Article

The rest of the article is organized as follows. All the notation used throughout the paper along with some preliminaries are presented in Section 3. In Section 4, an overview of the problem formulation and the overall system architecture is presented, while in Section 5 the search algorithm for finding the best planar cuts to fragment the mesh is analyzed. The generation of the dependency graph mechanism for identifying the precedence constraints between the chunks along with the generation of the manufacturing sequence of the chunks are analyzed in Section 6. A final post-processing step carried out to enhance the enhancing the adhesion between the chunks and guarantee the coverage of each chunk from its corresponding manufacturing path are presented in Section 7. The movement execution, control, disturbance estimation and the coordination of the UAV are presented in Section 8. Additionally, the hardware setup and the designed robotic platform along with the material selection can be found in Section 9. The whole framework evaluation and concept validation is verified in real-life experimentation results in Section 10 while a thorough discussion on them is carried out in Section 11. Finally, the article is concluded in Section 12.

## 3. Notations and Preliminaries

The notation and conventions described in the following will be used throughout the paper. All sets are notated with capital calligraphic letters as follows for set $\mathcal{S}$ and $|\mathcal{S}|$ denotes the cardinality (number of the items) of the set, while $\boldsymbol{\mathcal{S}}$ is a collection of sets $\mathcal{S}_i$. A three-dimensional vector in space is notated as $\vec{v} \in \mathbb{R}^3$.

A graph, denoted as $\mathbf{G} = \{\mathcal{V}, \mathcal{E}\}$, will be defined by the vertices set $\mathcal{V}$ and the edges set $\mathcal{E}$. A geometrical mesh $\mathbf{M} = \{\mathcal{V}, \mathcal{F}\}$ is considered a set of vertices $\mathcal{V} = \{v_i \mid i = 1, 2, \dots, N_M\}$ where $v_i = (x_i, y_i, z_i) \in \mathbb{R}^3$ and a set of faces $F = \{f_j \mid j = 1, 2, \dots, m\}$ where $f_j = (v_{i_1}, v_{i_2}, v_{i_3}) \mid v_{i_1}, v_{i_2}, v_{i_3} \in V$.

Additionally, three main frames are used throughout the paper and are shown in Fig.1. The world frame $\mathbb{W}$ is arbitrarily placed at the center of the room where the experiment is taking place, the UAVs body frame $\mathbb{B}$ which is placed on the center of mass of the UAV and the extruder's frame $\mathbb{E}$ which is placed at the tip of the extruder. The above notations and conventions will be applied consistently for clarity and precision.

## 4. System Overview

Initially, the original mesh $\mathbf{M}_{or}$ designed by the user in a Computer-Aided Design (CAD) software, with the intention of being constructed, is provided as input to the entire framework. The first step of the pipeline is to generate printing tasks for the available UAVs via splitting the mesh $\mathbf{M}_{or}$ into smaller sub-parts. This is handled by the Chunking algorithm, which searches for the best set of planar cuts $\mathcal{P}$ that will fragment the mesh into multiple chunks $C_i$ contained in set $\mathcal{C}$. Both executed cuts and the resulting chunks are handled in a structured way using a Binary Space Partitioning (BSP) Tree $\mathcal{T}$ [32] and each tree is given a heuristic score that facilitates their evaluation during the search algorithm. Upon completing the search and generating the final chunks, a dependency graph $\mathbf{G}$ is constructed. This graph comprehensively captures and manages all interdependencies among the chunks. This systematic approach provides a structured representation of the potential manufacturing pathways and sequences for reconstructing the original mesh. Furthermore, it enables the identification of available chunks for manufacturing at any given point in time. Through a dependency metric for each chunk in the graph, a manufacturing sequence $\mathcal{C}_{seq}$ is calculated iteratively. This sequence imposes the order in which the chunks will be manufactured sequentially from the available UAVs, ensuring an efficient manufacturing process. Additionally, after the chunks are generated, they undergo processing through a slicer software. This slicer after incorporating the corresponding hardware and material specifications of the mission, it transforms each chunk $C_k$ into a manufacturing path $\mathcal{P}_k$ which is to be traversed by the extruder while printing in order to successfully manufacture the chunk. A coordinator module is reactively commanding the UAV to continue to the printing process of the next chunk, as soon the currently assigned is completed or sending it back to its home position in case its battery is low. A schematic overview of the overall system architecture is shown in Fig. 2.

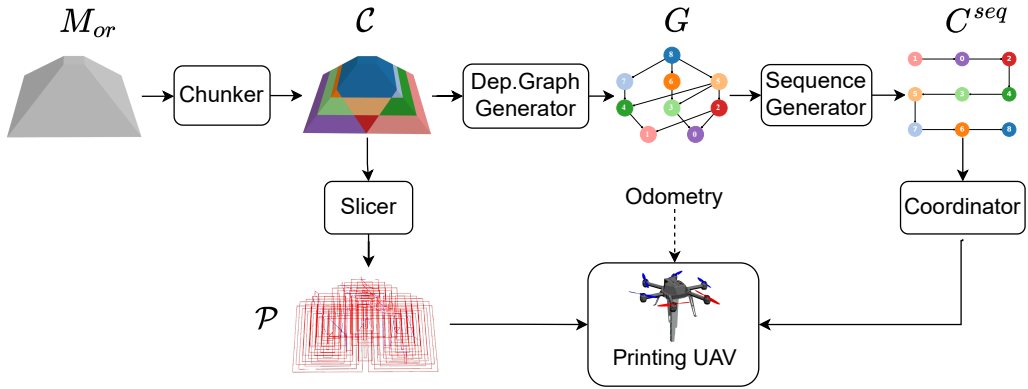

Figure 2: Schematic representation of the overall system architecture

The extrusion action along with the motion of the UAV are coordinated by an printing executor module. Specifically, the reference position of the UAV is fed to an onboard Offset-Free Non-Linear Model Predictive Controller (NMPC) that then interacts with the low-level flight controller leading to the desired path execution. The model used for the NMPC is dynamically updated using a disturbance estimator in order to improve the accuracy of the printing process. Simultaneously, direct servo commands are given to a servo controlling the extruder's valve and thus the flow of the construction material. A schematic overview of the UAV's onboard execution architecture is shown in Fig. 3.

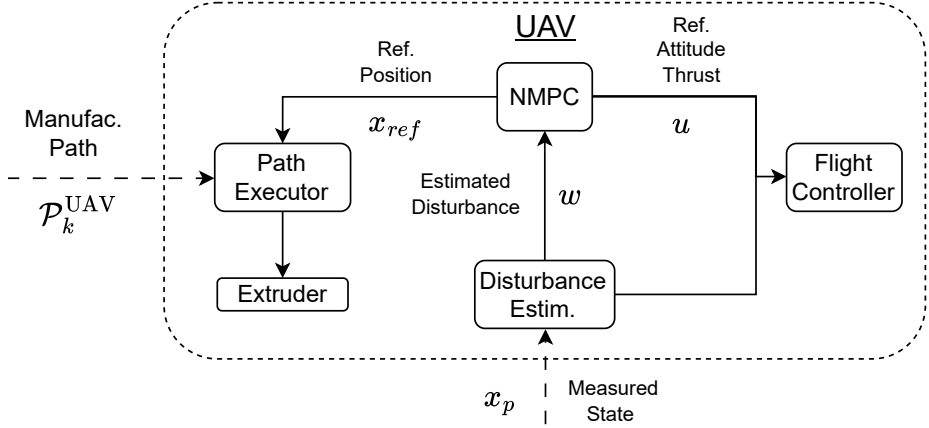

Figure 3: Printing UAV's onboard execution architecture

## 5. Mesh Decomposition - Chunking

The splitting of the given mesh is executed using multiple planar cuts that slice it into smaller sub-parts notated as chunks. Each planar cut, notated as $\Pi_k$, corresponds to a plane $p_{i,j}$ defined by the normal vector $\vec{n}_i \in \mathbb{R}^3$ and the origin point $\vec{p}_j \in \mathbb{R}^3$ which is a point laying on it. Whenever a cut $\Pi_k$ of a plane $p_{i,j}$ is executed with a single mesh $M_k$, it splits it into two parts, one laying over the plane in the half-space in the direction of the normal vector $\vec{n}_i$ which is called positive and notated as $M_k^+$ and the negative one $M_k^-$ that is laying in the half space below the plane in the opposite direction of the normal. All the sequential cuts to the original mesh $\mathbf{M}_{or}$ are handled via a BSP tree $\mathcal{T}$ where all of its leaves nodes correspond to the generated chunks $C_k$ and the planar cuts along with the resulting meshes at the various depths are stored in the intermediate ones. Whenever a new cut is executed, the affected chunk nodes are extended by following the convention of placing the negative chunk as the left subordinate. Thus, a systematic way of handling the chunks, executing new cuts, and extending the tree is realized.

The search for finding the best planar cuts set $\mathcal{P} = \{\Pi_0, \Pi_1, ..., \Pi_M\}$, where $M$ is the total number of selected cuts occurs iteratively, following a beam search algorithm [13]. This happens by generating a pool of normals $\vec{n}_i$ and uniformly sampling them from a spherical space $\mathcal{S}$. In each iteration of the algorithm, various possible extensions of the current search trees are evaluated by generating a family of planes for each sampled normal $\vec{n}_i$ by calculating origin points $\vec{p}_j$ along the direction of the normal. Between all the extensions of each tree $\mathcal{T}_\mu$, the top $W_{inner}$ are selected to be placed in the set of all the extended trees of the iteration. After the extensions of all the trees are placed in the set, the top $W_{outter}$ ones are selected to proceed to the next iteration. In this way, the computational burden of calculating all the possible trees , which scales exponentially in size and calculation time, is mitigated by forwarding the most promising trees to the following iterations. The search algorithm halts when all the $W_{outter}$ trees are considered terminal.

*5.1. Printability Constraints*

The sampling space $\mathcal{S}$ of the normals $\vec{n}_i$ is of radial distance $r = 1$. An upper bound, denoted as $\phi_{\max}$, is enforced on the polar angle $\phi$ to adhere to printing constraints, while the azimuthal angle $\theta$ is also constrained within $\theta_{\max}$. These bounds are carefully chosen based on the printability constraints and the limitations of the available hardware setup [13]. Specifically, to ensure sufficient overlap between adjacent faces of printed chunks, thus enabling consecutive layers to possess sufficient contact surface for bonding and resulting in a structurally robust outcome, an initial constraint $\phi_{\text{ar}}^{\max}$ is set on the polar angle $\phi$. Additionally, another bound $\phi_{max}^{coll}$ is imposed that ensures no collision between the extruder and the previously deposited material will occur. The extruder head is considered a rectangle of length $l$ and height $h$ resulting into $\phi_{max}^{coll} = arctan(h/l)$. Finally, the upper bound of the polar angle is considered the maximum of the aforementioned ones $\phi_{max} = max(\phi_{max}^{ar}, \phi_{max}^{coll})$ so finally $\phi \in [0, \phi_{max}]$. The constraint in the azimuthial angle $\theta$ is irrelevant of the hardware setup and is selected to be $\theta \in [-\pi, \pi]$ so that it covers the whole range.

*5.2. Tree Evaluation Heuristics*

In the evaluation process, each extended tree, denoted as $\mathcal{T}_i$, undergoes evaluation using a heuristic scoring system aimed at identifying the most promising candidates. This heuristic integrates various desired characteristics essential for optimal performance in the manufacturing process facilitated by the available UAVs. These attributes, defined mathematically, encompass qualities conducive to smooth, efficient and seamless manufacturing process. These features, are assessed collectively by either rewarding the favorable ones and penalizing the undesired ones.

*5.2.1. Volume Dispersion*

Uniform volume distribution among the chunks is preferred, so their manufacturing times will remain similar and lead to a smaller waiting time of the UAV being idle and waiting for the other to finish, thus resulting in a faster overall printing time. Furthermore, uniform volume distribution ensures an even arrangement of chunks, allowing for adequate spacing between their geometric centers. This trait is formulated using the volume dispersion metric $c_v = \sigma/\mu \in \mathbb{R}$ where $\mu = \sum_{i=1}^{N} V_i/N$, $\sigma = \sum_{i=1}^{N}(V_i - \mu)^2/N$ and $V_i$ denoting the volume of the $i$-th chunk, while $N$ the total number of the generated chunks. Finally, the resulting volume dispersion heuristic is as follows:

$$h_d = \mathcal{W}_d c_v \qquad (1)$$

*5.2.2. Seed Chunks*

Aiming to parallelize the generated printing tasks that are going to be assigned to the UAVs and thus make the whole manufacturing process faster, the number of seed chunks in the final tree is rewarded. Specifically, similarly to the notation in [33], a "seed" chunk is a chunk that all of its faces, apart from the one that is parallel to the ground, are acting as support to the adjacent chunks. With their presence, the manufacturing complexity is decreased since they can be printed on top of a flat chunk or on the ground and all of their faces act as supports for the other chunks. Towards the seed evaluation of a chunk $C_k$, its faces generated from the planar cuts are examined. A face $f_j$ that is co-planar with the cut $\Pi_i$ and lies on the negative half-space defined by it is called positive and is indicated by the term $f_j^+ = 1$. The necessary condition for the chunk to be considered seed is that all of its faces, excluding the one parallel to the ground, are positive. Its seed property is signified by the term $s_k = 1$. An indicative example of a seed chunk $C_k$ is shown in Fig. 4. There are three planes $\Pi_i, \Pi_j, \Pi_h$ involved in the generation of its faces and the two of them are defining cuts in which the mesh is in the negative side of both of them while the $\Pi_h$ is a ground plane and is not taken into consideration in the evaluation.

The seed chunk is rewarded by including it in the heuristic $g(C_k)$ for any chunk $C_k$ while the number of the positive faces is included as well as shown in Eq. 2.

$$g(C_k) = \mathcal{W}_s s_k + \mathcal{W}_f \sum_{j=0}^{\Gamma} f_j^+ \qquad (2)$$

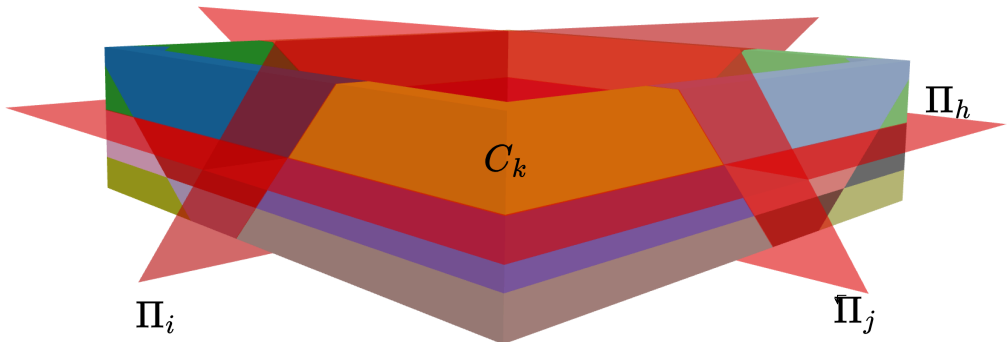

Figure 4: Seed Chunk $C_k$ Representation with Faces Touching Planes: Faces for $\Pi_i$ and $\Pi_j$ are Positive, while $\Pi_h$ is Parallel to the Ground.

where $\mathcal{W}_s, \mathcal{W}_f \in \mathbb{R}^-$ are tunable weights. In this particular case, since this term needs to be rewarded the gains are selected to be non-positive.

*5.2.3. Ground planes*

It is observed that planar cuts with normals perpendicular to the ground facilitate the construction process. This is because the resulting chunks, upon extraction, rest neatly atop one another, providing sufficient area for support. Moreover, the flat faces of these chunks contribute to enhancing the quality of the final construction. Despite potential inaccuracies in UAV and extruder placement, these flat faces offer stable surfaces to support subsequent layers. Therefore, the selection of such cuts is considered advantageous for the process. However, it is not mandatory to include them in every instance, as an excess of ground cuts can be unfavorable since too many ground cuts increase dependencies and fail to parallelize printing tasks efficiently. Thus aiming to let the search algorithm select them, the number of ground cuts is rewarded and included in the heuristic function. A planar cut $\Pi_i$ is considered a ground cut if its normal vector $\vec{n}_i$ is parallel to the ground plane normal $\vec{n}_g = [0, 0, 1]$. So the term $g_i$ referring to the planar cut $\Pi_i$ is included in a sub-heuristic $h_g$

$$h_g = \sum_{i=0}^{M} g_i \quad (3)$$

where $M$ corresponds to the number of planar cuts and

$$g_i = \begin{cases} 1 & \text{, if } \vec{n}_i \times \vec{n}_g = 0 \\ 0 & \text{, otherwise} \end{cases} \quad (4)$$

All the planes of the set $\mathcal{P}$ identified as ground ones as well, are placed in the subset $\mathcal{P}_G \subseteq \mathcal{P}$.

*5.2.4. Critical size Chunks*

Throughout the execution of multiple planar cuts, it is possible that small chunks may appear. These are considered side-effects of the chunking algorithm and are not useful for the overall manufacturing of the original mesh since they result in extra movements and waiting time for the rest of the UAVs. Aiming to mitigate such kind of chunks, a critical threshold $V_{cr} \in \mathbb{R}$ is considered on their volume. The existence of chunks having a volume less than the threshold is penalized by incorporating into the heuristics as follows:

$$h_V^{cr} = \sum_{i=0}^{N} \mathcal{W}_{cr} s_k^{cr} \tag{5}$$

where

$$s_k^{cr} = \begin{cases} V_k & \text{, if } V_k < V_{cr} \\ 0 & \text{, otherwise} \end{cases} \tag{6}$$

Since the volume is penalized only if it crosses the lower bound and $\mathcal{W}_{cr}$, a tunable gain set to a high value represents an impulse penalization of the small chunks.

In similar fashion. chunks with at least one dimension smaller than the printing line width are likely to cause issues during the slicing process, where manufacturing paths are generated. These chunks may not be identified correctly or, due to their small size, may make execution more difficult. Following that, each chunk is further evaluated by generating the Oriented Bounding Box(OBB) $\mathcal{O}_{BB}$ corresponding to it. The OBB is a generalized form of a simple bounding box that is also rotated in the 3D space so that it encaptures the given mesh completely. It is defined as follows:

$$\mathcal{O}_{\text{BB}} = \{p \in \mathbb{R}^3 : p = C + \lambda_1 \vec{u}_1 + \lambda_2 \vec{u}_2 + \lambda_3 \vec{u}_3\} \tag{7}$$

where $C \in \mathbb{R}^3$ denotes the center of the box, $\vec{u}_1, \vec{u}_2, \vec{u}_3$ are orthonormal unit vectors defining the orientation of the box and $e_1, e_2, e_3 \in \mathbb{R}$ are the extents along each axis. In the case that the minimum of the extents $e_{\min} = \min(e_1, e_2, e_3)$ is less than the line-width $l_w$, another critical penalization occurs. It is formally formulated as follows:

$$h^{cr}_{\text{BB}} = \sum_{i=0}^{N} \mathcal{W}_{cr} b^{cr}_k \tag{8}$$

where

$$b^{cr}_k = \begin{cases} 1 & \text{, if } e^k_{\min} < l_w \\ 0 & \text{, otherwise} \end{cases} \tag{9}$$

Finally the two aforementioned heuristic of critical size and dimension penalization are fused into a common one:

$$h_{cr} = h^{cr}_V + h^{cr}_{\text{BB}} \tag{10}$$

*5.2.5. Overall Heuristic*

All the aforementioned heuristics for any given BSP tree $\mathcal{T}$ are fused into a common one $h_\mathcal{T} \in \mathbb{R}$ by summing them as follows:

$$h_\mathcal{T} = h_d + \sum_{k=0}^{N} g(C_k) + h_g + h_V \tag{11}$$

*5.3. Terminal Condition*

Making the chunking adaptive to the available UAV fleet characteristics, the beam search is conducted iteratively until all outer trees $\mathcal{W}_{outter}$ forwarded to the next search iteration are terminated. A tree $\mathcal{T}$ is deemed terminal when all of its components can be printed by at least one UAV. This follows the convention that each chunk is printed by only one UAV, and one UAV can print multiple chunks, but not vice versa. Towards that, the volume of the material carried by a UAV $r$ is denoted as $d_r$ and all the available UAVs' volumes are placed in a descending order in the set $\mathcal{D}$. Similarly, the volume of the chunk $C_k$ is denoted as $c_k$ and all of them are placed in descending order in the set $\mathcal{C}$. Thus, the tree is considered terminated when a non-empty set $\mathcal{S}$ exists and is defined as follows:

$$S = \{a_l = (d_j, c_i) : d_j > c_i\} \neq \emptyset, \forall l = [1, ..., k] \tag{12}$$

and the aforementioned sets are updated recursively.

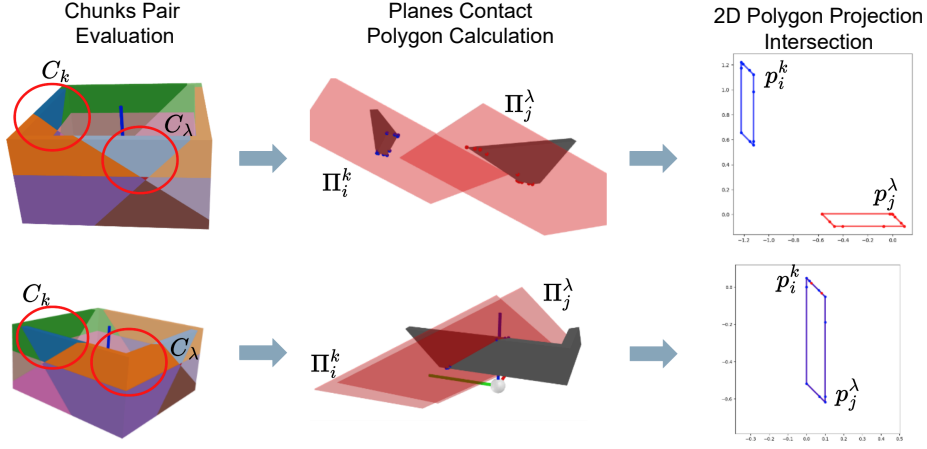

Figure 5: Plane contact and face contact checking during Dependency Graph Generation. Depsite the cooorespondling planes of faces are touching in both cases, there is contact on the faces level only on the bottom one.

## 6. Dependency Graph - Printing Tasks Assignment

After having generated all the chunks, there are various ways the manufacturing process could be conducted so that the original mesh is constructed. The constraint that the negative chunks should be printed prior to the positive ones must be followed since they act as support for them. Aiming to capture all their inter-dependencies, a dependency graph is calculated that transforms the BSP tree $\mathcal{T}$ to a graph mechanism used to indicate the chunks that can be printed at any given time without violating the rule mentioned before and be updated recursively whenever a new chunk is manufactured. Specifically, a directed graph $\mathbf{G} = (\mathcal{C}, \mathcal{E})$ is derived from the generated BSP tree, where the set of nodes is denoted as $\mathcal{C} = \{C_k, \ i = 0, .., N\}$ corresponding to the set of generated chunks and $N$ equal to the number of generated chunks and the set $\mathcal{E} = \{E_{i,j}\}$ containing all the directed edges. An edge $E_{i,j}$ between nodes representing chunks $C_i$ and $C_j$ exists if and only if $C_i$ is dependent on $C_j$. An algorithmic approach is conducted to calculate the dependency between two chunks.

Toward the calculation of dependency between two chunks $C_k$ and $C_\lambda$, all of their faces are evaluated in pairs. Initially, all planar cuts $\Pi_i$ of the tree that were involved in the generation of chunk $C_k$ are gathered in a set $\mathcal{P}_k$ and all the faces of the chunk are placed into the set $\mathcal{F}_k^{all} = \{f_i^k, i = 0, 1, M_i\}$ where $f_i^k$ denotes the plane corresponding to the $i$-th facet of the chunk $C_k$.

The intersection of the aforementioned sets $\mathcal{F}_k = \mathcal{P}_k \cap F_k^{all}$ is calculated and used for the remainder of the process. Having computed the sets $\mathcal{F}_k$ and $\mathcal{F}_\lambda$ for the chunks, the pairs of their faces are evaluated for dependency between them. This happens in a two-step method, initially, a check is carried out for evaluating whether the two planes corresponding to the faces are co-planar, and if this is true the contact between the faces is calculated. The condition to determine that the planes $\Pi_i^k, \Pi_j^\lambda$ of two facets $f_i^k, f_j^\lambda$ are co-planar is having their normal vectors $\vec{n}_i$ and $\vec{n}_j$ parallel to each other while at the same time the distance between them $d_{\lambda,j}^{k,i}$ is less or equal than a low positive threshold $\epsilon \in \mathbb{R}^+$. The distance between planes is defined as follows:

$$d_{\lambda,j}^{k,i} = |(\vec{o_i^k} - \vec{o_j^\lambda}) \cdot \vec{n_i}| \tag{13}$$

where $\vec{o_i^k}, \vec{o_j^\lambda}$ are points lying on the planes of faces $f_i^k$ and $f_j^\lambda$ and $\vec{n}_i$ the normal vector corresponding to both of them since it has already been verified that they are parallel in the previous check. Given that the condition $d_{\lambda,j}^{k,i} \leq \epsilon$ is true (where $\epsilon \in \mathbb{R}$ is a small number), the planes of the faces are in contact but the faces themselves need to be checked further to verify the dependency. Towards this examination, the polygons $p_i^k$ and $p_j^\lambda$ generated from the edges of the faces $f_i^k$ and $f_j^\lambda$ are calculated and projected into the 2D plane defined by the faces' planes. The calculation of the overlapping area between the 2D polygons is carried out and in case it is greater than zero it is signified that the two chunks are in contact and concluding in the fact that there is a dependency relation between them. A visual representation of the steps carried out toward the evaluation of the contact between two pairs of chunks is shown in Fig. 5.

On determining the dependency between chunks, the constraint that the vertical $z$ component of the normal vector $\vec{n}_i$ is consistently positive, ensuring the vector always points upwards, is leveraged. Specifically, the characteristic of positive and negative chunks, as detailed in Section 5, is exploited to identify the dependent chunk. Given that the positive chunk $C^+$ consistently resides in the half-space defined by the normal vector's direction, it inevitably lays on top of the negative one $C^-$, establishing that the positive is dependent on the negative. The pseudocode for the dependency calculation process is shown in Algorithm 1.

**Algorithm 1** Dependency Calculation of $C_k$ to $C_\lambda$
---
**for** $f_k^i$ in $\mathcal{F}_i$ **do**
    **for** $f_\lambda^j$ in $\mathcal{F}_j$ **do**
        **if** $f_k^i$.parallel($f_k^j$) and $f_k^i$.distance($f_k^j$) $\epsilon$ **then**
            $pol_i^k \leftarrow getPolygon(f_k^i)$
            $pol_j^\lambda \leftarrow getPolygon(f_k^j)$
            $area \leftarrow getOverlap(pol_i^k, pol_j^\lambda)$
            **if** $area \geq 0$ and $f_k^i.negativeTo(f_k^j)$ **then**
                return **True**
            **end if**
        **end if**
    **end for**
**end for**
return **False**
---

*6.1. Printing Sequence*

Each chunk $C_k$ is considered as an individual printing task that is going to be assigned to one of the available UAVs. Having computed the graph **G**, a sequence $\mathcal{S}$ in which the chunks are going to be manufactured is selected. This sequence needs to be with respect to the geometrical constraints of the chunks, as encaptured by the dependency graph while at the same time, the aim of alleviating the graph from complex dependency schemes is favorable. The graph serves as a mechanism that contributes to the discovery of the available chunks that can be printed at any given time since the state of the chunks can be set to manufactured and are updated dynamically. A chunk $C_k$ is considered available to be printed if and only if no nodes are dependent on it or if all of its dependent nodes have been manufactured. All the manufactured nodes are placed in the set $\mathcal{M}$. Thus, the availability of a chunk can be mathematically formulated using the term $a_k = 1$ which is defined as follows:

$$a_k = \begin{cases} 1, & \text{if } |\mathcal{N}^-(C_k)| = 0 \text{ or } \mathcal{N}^-(C_k) \subseteq \mathcal{M} \\ 0, & \text{otherwise} \end{cases} \quad (14)$$

where $\mathcal{N}^-(C_k)$ denotes the set of all the descendants of the node corresponding to the chunk $C_k$ and $|.|$ represents the cardinality of the set. At any given moment, the chunks that can be printed are gathered in the set $\mathcal{A} = \{C_i : a_i = 1, \forall i = 1, 2, \ldots N_a\}$. where $N_a$ corresponds to the total number of the available chunks. In the realm of simplifying the complexity of the printing sequence, the chunks with the most dependent nodes on them are prioritized in the selection process for generating the manufacturing sequence. Therefore, a score $s_k = |\mathcal{N}^+(C_k)|$ is assigned to each chunk node where $\mathcal{N}^+(C_k)$ denotes the set of all the predecessors of the node corresponding to chunk $C_k$. On top of that, a manufacturability constraint is imposed regarding the relative position of the chunks with respect to the utilized ground planar cuts $\Pi_i \in \mathcal{P}_G, i = 0, 1, \ldots, N_G$ where $N_G$ denotes their total number. Particularly, $N_G + 1$ number of chunk layers $\mathcal{L}_i^G, i \in [1, N_G+1]$ are formed, each one of them bounded between either the ground and a ground plane or two consecutive ground planes. An iterative process is carried out where in each iteration the best chunk from the set $\mathcal{A}_\mathcal{G}$ is evaluated and selected to be the next one to be assigned and manufactured to the available UAV. The set $\mathcal{A}_\mathcal{G}^i = A \cap \mathcal{L}_i^G$ is defined as the intersection of the sets of available chunks to be printed and the chunks of the ground layer $\mathcal{L}_i^G$. Initially, $i$ is set to 0 and it is increased by one whenever all the chunks of the layer contained in the set $\mathcal{L}_i^G$ are manufactured, hence $\mathcal{A}_\mathcal{G}^i = \emptyset$. As soon as a chunk is assigned, it is placed in the set $\mathcal{M}$ along with all the manufactured ones and the graph $\mathbf{G}$ is updated. The iterative process is carried out until $\mathcal{A} = \emptyset$, indicating that no more chunks are available and all of them have been manufactured, resulting in the construction of the given mesh. An overview of the printing sequence calculation process is shown in Algorithm 2.

*6.2. Chunks Assignment*

The assignment of the chunks to each material canister is executed based on the volume of it. Specifically, each canister is used only once by the robot, so all of canister volumes $d_i$ are placed in the set $\mathcal{D}$ in a descending order. An iterative process is carried out where the chunks of the sequence $\mathcal{S}$ are assigned to each canister $d_i$ and the assignments are placed in the set $\mathcal{A}_{as}$ as follows $\mathcal{A}_{as} = \{(c_i, d_j) | c_i \in \mathcal{C}, d_j \in \mathcal{D}\}$ until the available material can no longer accommodate for the next chunk in $\mathcal{S}$. In that case, the canister is considered fully assigned and the algorithm proceeds to the next one while $\mathcal{S}$ is updated. The iterative process is carried out until there are no chunks left in $\mathcal{S}$. An overview of the aforementioned process in pseudocode can be found in Algorithm 3.

**Algorithm 2** Printing Sequence Calculation
---
    $i \leftarrow 0$
    **while** $\mathcal{A} \neq \emptyset$ **do**
        $\mathcal{A} \leftarrow getAvailableNodes(G, \mathcal{M})$
        $\mathcal{L}_G^i \leftarrow getChunksOfLayer(G, i)$
        $\mathcal{A}_G \leftarrow interesection(\mathcal{A}, \mathcal{L}_G^i)$
        **if** $\mathcal{A}_G == \emptyset$ **then**
            $i \leftarrow i + 1$
        **else**
            $C_{nxt} \leftarrow getTopChunk(\mathcal{A}_G)$
            $assignNextChunk(C_{nxt})$
            $\mathcal{M}.append(C_{nxt})$
        **end if**
    **end while**

**Algorithm 3** Chunks Assignment to Canisters
---
**Require:** $\mathcal{S} = \{s_1, s_2, ..., s_n\}$
**Require:** $\mathcal{D} = \{d_1, d_2, ..., d_m\}$       ▷ Sorted in descending order
    $\mathcal{A}_{as} \leftarrow \emptyset$
    $i, j \leftarrow 1, 1$       ▷ Canister, Chunk index
    **while** $i \leq n$ **do**
        $v_i \leftarrow d_i$       ▷ Remaining volume of canister $i$
        **while** $j \leq n$ and $s_j \leq v_i$ **do**
            $v_i \leftarrow v_i - s_j$       ▷ Update rem. volume of canister
            $\mathcal{A}_{as} \leftarrow \mathcal{A}_{as} \cup \{(s_j, d_i)\}$       ▷ Update assignments
            $j \leftarrow j + 1$
        **end while**
        $i \leftarrow i + 1$       ▷ Move to the next canister
    **end while**
    **Return** $\mathcal{A}_{as}$       ▷ Return the final assignment list

## 7. Chunks Interlocking Mechanism

After all chunks are generated and the dependency graph is calculated as stated in both Sections 5 and 6, a post-processing step is carried out in order to reinforce the cohesiveness between chunks in contact and ensure that the manufacturing paths are covering each chunk entirely, without any gaps. Specifically, after the chunking, the two faces of the contacting chunks generated by the planar cut are responsible for their interconnection. However, given the fact that the layer height is comparable to the overall dimensions of the chunks (not a usual phenomenon in conventional 3D printing), there are cases during the segmentation of the chunk into layers from the slicing software where regions of each layer will have a lower height due to the inclination of the planar face. These areas are not sliced since they do not suffice to be considered as part of the chunk to be manufactured and are discarded. Examples of such regions are shown in Fig. 6.a annotated with green color.

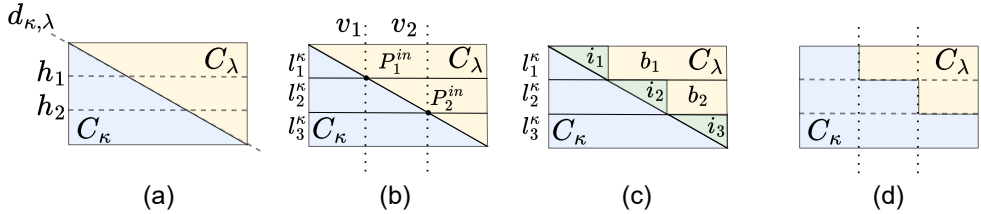

Figure 6: Chunks Interlocking Process Intermediate steps. The mesh is splitted into layer segments (a), vertical lines are calculated in the intersection of layers and the faces (b), the regions of inclination are calculated (c) and the final "stair-case" pattern is formed.

Aiming to accommodate for such resolution artifacts, an intermediate process is introduced between the chunking and the final slicing of all chunks $C_k \in \mathcal{C}$. Initially, all pairs of chunks in contact are placed in a set

$$\mathcal{C}^{con} = \{(C_k, C_l) : f_c(C_k, C_l) = 1\} \quad (15)$$

where $f_c(C_k, C_l)$ signifies the contact between the $k$-th and $l$-th chunks and a connection algorithm is carried out for each pair in it. Both chunks are explicitly sliced into layer segments $l_i^k \in \mathcal{L}^k, l_j^\lambda \in \mathcal{L}^\lambda$ by cutting them with a vertical plane in intervals equal to the layer height $l_h \in \mathbb{R}$ (shown in Fig. 6.b). Later, a calculation of the points $P_\mu^{in}$ is carried out, where $P_\mu^{in}$ are the intersection points between the diagonal line $d_{\kappa,\lambda}$ defined by the faces in contact and the horizontal lines $h_i$ used to segment the chunks in layers. For every point $P_\mu^{in}$, a vertical line $v_i^{\kappa,\lambda} \in \mathcal{V}^{\kappa,\lambda}$ is passing through it, separating the top chunk $C_\lambda$ further into regions of inclination $i_i \in \mathcal{I}^{\kappa,\lambda}$ and regions with flat borders $b_i \in \mathcal{B}^{\kappa,\lambda}$ as shown in Fig. 6.c. Finally, a convention is used defining that during the interlocking process, the artifacts due to layer segmentation are incorporated into the bottom chunk of each pair in contact and extend it so that they form a staircase-like pattern. The green regions notated with $i_i$ are connected with their corresponding layer segment $l_i^\kappa$ resulting into the final result shown in Fig. 6.d. An overview of the whole process is shown in Algorithm 4.

**Algorithm 4** Chunks Interlocking Algorithm
---
$\mathcal{C}^{con} \leftarrow getChunkPairsInContact()$
**for** $C_i, C_j$ in $\mathcal{C}^{con}$ **do**
    $\mathcal{L}^k, \mathcal{L}^\lambda \leftarrow sliceHorizontalLayers(C_i, C_j)$
    $\mathcal{V}^{\kappa,\lambda} \leftarrow calculateVertPlanes(\mathcal{L}^k, \mathcal{L}^\lambda, d_{\kappa,\lambda})$
    $\mathcal{I}^{\kappa,\lambda}, \mathcal{B}^{\kappa,\lambda} \leftarrow sliceLayersWithPlanes(\mathcal{L}^\lambda, \mathcal{V}^{\kappa,\lambda})$
    $\mathcal{L}^k \leftarrow connectFragmentsToBottom(\mathcal{L}^k, \mathcal{I}^{\kappa,\lambda})$
    $C_i, C_j \leftarrow uniteBackTogether(\mathcal{L}^k, \mathcal{B}^{\kappa,\lambda})$
**end for**

The aforementioned process steps are shown in 3D in Fig. 7, where two chunks in contact originated from a decomposed solid rectangle are connected using the staircase pattern. Initially, the vertical planes are calculated and visualised (top), later along with the horizontal plane the parts $i_1, i_2 \in \mathbb{I}$ are calculated after the slicing with the planes and are visualised with green (middle). The final processed chunks are shown (bottom) where the "staircase" pattern between the chunks is formed. All the chunks of the solid box before and after the interlocking post-processing module are shown in Fig. 8.

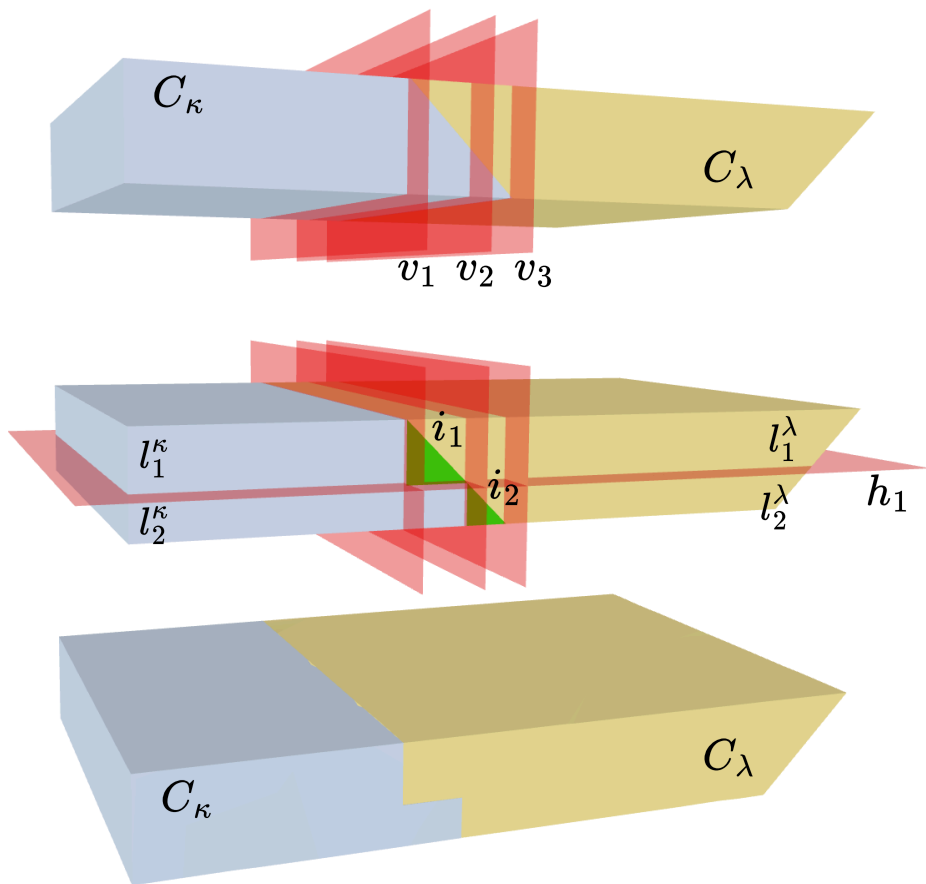

Figure 7: 3D visualization of the intermediate steps during during the interlocking process of two chunks (blue and yellow) in contact generated from a decomposed solid rectangle.

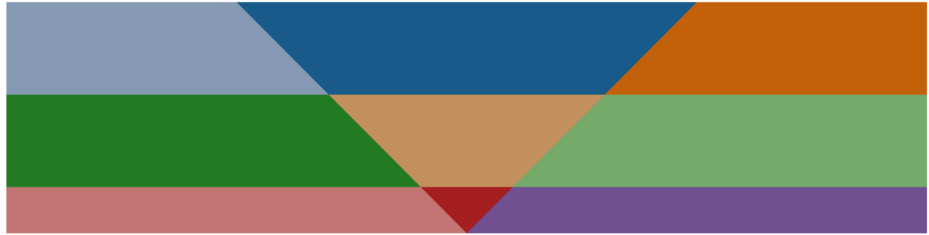
(a) Chunks before the interlocking process

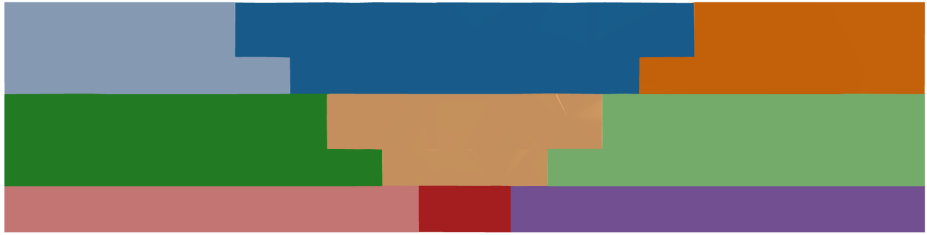
(b) Chunks after the interlocking process

Figure 8: Chunk Interlocking Mechanism applied to a decomposed mesh: Initially, planar cuts define the connections (a), but after processing (b), increased contact areas improve adhesion and ensure proper handling during slicing.

## 8. UAVs Printing Execution

Before any chunk $C_k$ is assigned to an available UAV, a series of steps is performed so that the 3D geometry of the mesh is transformed into actual movements of the UAV and deposition of the carried material. Specifically, a slicer software accommodates for the 3D triangles set to manufacturing paths that the UAV needs to traverse in order to achieve the manufacturing of the respective part. Having generated the path $\mathcal{P}_k$ for the UAV, it is passed to the printing executor module that coordinates the motion of the UAV by interfacing a Non-Linear Model Predictive Controller (NMPC) and directly controlling the extrusion of the material.

### 8.1. Chunks Slicing - Manufacturing Paths

Each chunk $C_k$ is handled in a stereo-lithography format(.stl) consisting of a set of vertices and triangles connecting them. The transformation of this abstract geometric structure to meaningful paths that the UAV will traverse is handled by an open-source slicer software 'Cura' [34] which is considered one of the standards for classic 3D printing. Specifically, adjusting the parameters and setting up the software in the particularities of the aerial 3D printing setup such as the material line width, layer height, extruder dimensions, etc., the path $\mathcal{P}_k^{ex}$ is generated. This 3D path is consisted of points that the extruder needs to visit by executing linear movements in each segment so that it results in the construction of the given chunk. Additional information generated from the slicer such as extruder temperature, moving speed and extrusion rate are neglected since there is no need for a heating element for the current material section and both UAV speed and extrusion rate of the material are handled by the printing executor module. However, the path $\mathcal{P}_k^{ex}$ corresponds to the extruder movement, not the UAV itself. Assuming that the roll and pitch angle of the UAV is sufficiently small so that its rotation does not diverge the tip of the extruder, a vertical elevation of the path to match the body frame of the UAV is performed and the final path $\mathcal{P}_k^{\text{UAV}}$ is calculated. The generated manufacturing paths for two chunks that are later going to be assigned to the UAVs are shown in Fig. 9.

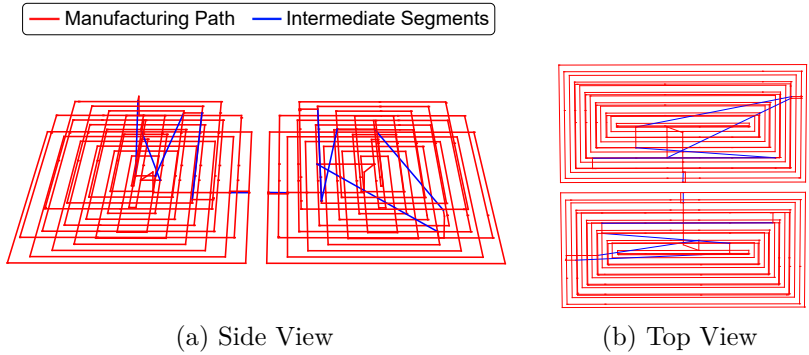

(a) Side View        (b) Top View

Figure 9: Generated Manufacturing paths of two chunks after slicing. The Manufacturing path is marked with red while the intermediate segments where no material extrusion occurs are marked with blue.

*8.2. Printing executor*

Once the path $\mathcal{P}_k^{UAV}$ is generated, it is sent to the printing executor module. This module manages all the movements of the UAV, and it is responsible for coordinating with the chunker to receive the generated chunks. At the same time, it interfaces with the extrusion mechanism to control the material deposition. Additionally, it provides position references to the NMPC module, which handles the traversal of the manufacturing path. This is done by assuming a constant deposition speed $V_{dep} \in \mathbb{R}$ during the printing execution. Each linear segment $l_i$ of the path $\mathcal{P}_k^{UAV}$ is interpolated and the sequential points are provided to the NMPC with respect to the aforementioned speed $V_{dep}$. In parallel with the traversal of the path, the executor commands the start of the deposition of the material right before the execution of the first linear segment and commands its halt at the end of the traversal since various movements are occurring from the UAVs' home positions to the printing workspace.

*8.3. Non-Linear Model Predictive Controller (NMPC)*

Concerning the movement of the UAV in space, an onboard Non-Linear Model Predictive Controller (NMPC) [35] is utilized. As mentioned before, it is commanded by the printing executor that is feeding it position references for the UAV and it executes the movements by interfacing with an onboard flight controller. Specifically, the UAV dynamics are described via the world frame $\mathbb{W} \in \mathbb{R}^3$ and the body frame $\mathbb{B} \in \mathbb{R}^3$ attached to the center of mass of the UAV as shown in Fig. 1.

The dynamic non-linear equations governing the motion of the UAV in the 3D space are as follows:

$$\begin{aligned}
\dot{p}(t) &= v(t) + w_p \\
\dot{v}(t) &= R_{x,y}(\theta, \varphi)\mathbf{T} + \mathbf{G} - \mathbf{A}v(t) + w_v \\
\dot{\varphi}(t) &= \frac{1}{\tau_\varphi}(K_\varphi \varphi_{\text{ref}}(t) - \varphi(t)) + w_\varphi \\
\dot{\theta}(t) &= \frac{1}{\tau_\theta}(K_\theta \theta_{\text{ref}}(t) - \theta(t)) + w_\theta
\end{aligned} \qquad (16)$$

where the position of the UAV in any given time $t$ is denoted by $p(t) \in \mathbb{R}^3$ and the velocity by $v(t) \in \mathbb{R}^3$ referring both to the world frame $\mathbb{W}$. The roll, pitch and yaw angle of the UAV are denoted by $\varphi, \vartheta, \psi \in \mathbb{R}$ respectively while $R(\varphi, \theta, \psi) \in SO(3)$ is the rotation matrix representing the transformation of thrust-induced acceleration from the UAV's body frame to the corresponding components in the world frame. The attitude dynamics are modeled via first-order response systems of time constants $\tau_\varphi$ and $\tau_\vartheta$. They are incorporated in this way since a low-level onboard controller is being executed in the flight controller of the UAV and its tracking the reference angles $\varphi_{ref}, \vartheta_{ref} \in \mathbb{R}$ and the reference thrust value $T_{ref} \in \mathbb{R}$ similar to [35]. These three terms form the vector $u = [\varphi_{ref}, \theta_{ref}, T_{ref}]$ that is considered as control variable of the control loop. Finally, terms $w_p, w_v \in \mathbb{R}^3$ and $w_\varphi, w_\theta \in \mathbb{R}$ are velocity, force and angular roll and pitch rate disturbances constituting the disturbance vector $w = [w_p, w_v, w_\varphi, w_\theta] \in \mathbb{R}^8$ which is later analyzed in Section 8.4. The non-linear continuous model of Eq. 16 is discretized using the Forward Euler Method through a sampling time of $\delta_t \in \mathbb{R}$ within a prediction horizon of $N$ timesteps and contributes in the prediction of the state of the UAV in each time instant $(k+j|k)$, where $(k+j|k)$ denotes the prediction of the time step $k+j$ computed at the time step $k$. The objective of the controller is to compute an optimal sequence of input vectors $u$ within the horizon time steps in order to minimize the cost function $J$ which is defined as follows:

$$J = \sum_{j=1}^{N} \underbrace{(x_{ref} - x_{k+j|k})^T Q_x (x_{ref} - x_{k+j|k})}_{\text{state deviation cost}}$$
$$+ \underbrace{(u_{nom} - u_{k+j|k})^T Q_u (u_{nom} - u_{k+j|k})}_{\text{nominal input cost}} \quad (17)$$
$$+ \underbrace{(u_{k+j|k} - u_{k+j-1|k})^T Q_{\Delta u} (u_{k+j|k} - u_{k+j-1|k})}_{\text{input smoothness cost}}$$

where the state vector is denoted by $x = \{p_x, p_y, p_z, v_x, v_y, v_z, \varphi, \vartheta\}$, $Q_x \in \mathbb{R}^{8 \times 8}$, $Q_u$ and $Q_{\Delta_u} \in \mathbb{R}^{3 \times 3}$. The deviation of the predicted future states from the reference one is incorporated into the "state deviation cost". Towards aiming to minimize the energy of the control action of the controller a nominal input $u_{nom} = \{0, 0, g\}$ is defined corresponding to the state that the UAV is hovering stationary. Any deviations from this input during the operation are penalized using the "nominal input cost" term. Any differences between sequential inputs are penalized via the "input smoothness cost" term to enforce smooth control actions since large discontinuities of the control vector between time steps are proportionally penalized.

The input vector $u$ is bounded by the lower and upper bounds $u_{min}, u_{max} \in \mathbb{R}^3$ while additional constraints are imposed to the rate of the absolute rate of change of reference angles $\varphi_{ref}, \vartheta_{ref}$ denoted with the bounds $\Delta\varphi_{max}$ and $\Delta\theta_{max}$. On top of the aforementioned inequality constraints, the model dynamics of Eq. 16 are formulated as equality constraints and used as the basis of the prediction part of the optimization. The non-linear optimization problem is solved through the Optimization Engine (OpEn) software [36] in which the PANOC method with a single-shooting approach [37] is utilized.

The yaw angle $\psi$ of the UAV is not considered in the model predictive controller but is controlled using a decoupled PID controller sending direct yaw rate commands $\dot{\psi}_{ref}$ to the onboard low-level flight controller.

*8.4. Offset-Free MPC - Non-Linear Moving Horizon Estimator (NMHE)*

The aforementioned MPC provides smooth and responsive flight, however, a steady state offset during tracking of the given paths is present. This constant deviation from the goal position can be accounted to various causes from unbalanced manufacturing of the UAV, miscalibrated sensors, actuator mismatch, inaccurate model used in the prediction scheme, ground effect and gradual weight loss due to the extrusion of the material. Given the fact that the application of aerial additive manufacturing requires meticulous precision and accuracy in order to manufacture the given mesh, these offsets need to be eliminated since any deviation from the path will result in the incorrect placement of the chunk that is being printed and consequently, may cause problems to the assembly of the whole structure. Given the fact that these inaccuracies and errors can be limited but not fully eliminated, similar to the methodology of the offset-free MPC [38, 39] all the aforementioned deficiencies are modeled as disturbances to the initial nominal model and are estimated in real-time while the UAV is airborne. In particular, the UAV is considered as a time-invariant dynamical system in the form of

$$\begin{aligned} \dot{x}_p &= f_p(x_p, u) \\ y_p &= h(x_p, u) \end{aligned} \quad (18)$$

where $x_p \in \mathbb{R}^8$, $u \in \mathbb{R}^3$ and $y_p \in \mathbb{R}^8$ denote the current plant state, the input and the output of the plant. The functions $f_p(x_p, u) : \mathbb{R}^8 \times \mathbb{R}^3 \times \mathbb{R}^8 \to \mathbb{R}^8$ and $h_p(x_p) : \mathbb{R}^8 \to \mathbb{R}^8$ are not known precisely and are assumed to be continuous.

The model used for the prediction scheme of the MPC is defined as the nominal and is as follows:

$$\begin{aligned} \dot{x} &= f(x, u) \\ y &= h(x) \end{aligned} \quad (19)$$

where $x \in \mathbb{R}^8$ and $y \in \mathbb{R}^8$ denote the current state and the output of the model and the dynamics function $f(x, u)$ correspond to the nominal model dynamics of the UAV, initially introduced in Eq. 16. Thus, the deviation between the plant and the model $w \in \mathbb{R}^8$ is formulated as shown below:

$$w = f_p(x_p, u) - f(x, u) \quad (20)$$

In this specific application, the state of the plant is fully measurable, which simplifies both $h_p(x_p, v_p)$ and $h(x)$ to $x_p$ and $x$, respectively. Furthermore, it is assumed that the state disturbance is bounded within a compact set $w \in \mathbb{W}$, and that this disturbance is additive in nature. Its value changes over time during the plant's operation, so it must be estimated in real time while the UAV is flying.

Towards that, a Nonlinear Moving Horizon State Estimation (NMHE) approach is utilized to estimate the disturbance $w$ in the aforementioned model. A window of past measurements of both state and control input is kept locally and an optimization problem is formulated aiming to estimate a value for $w$ which minimizes the deviation of the model estimation from the actually measured states. Each estimation window is of size $N_e$ and the disturbance is assumed to be static throughout it leading to $\dot{w} = 0$. Following the aforementioned assumptions, the estimation model is formulated as follows:

$$\begin{aligned} \dot{x} &= f(x, u) + w \\ \dot{w} &= 0 \\ y &= x \end{aligned} \quad (21)$$

Similarly to the model of the MPC mentioned in section 8.3, the model is discretized using the Forward Euler Method via a sampling time of $\delta_t \in \mathbb{R}$ within an estimation window of $N_e$ time-steps and contributes in the prediction of the state of the UAV in each time instant. At each time instant $k$ throughout the whole mission, a set $Y = \{y_{k-i}, i = 0, \ldots, N_e\}$ of the previous state measurements and a set $U = \{u_{k-i}, i = 0, \ldots, N_e\}$ of the previous inputs are gathered in order to be used for the estimation. Specifically, a set of state estimates $\hat{X} = \{\hat{x}_{k-i}, i = 0, \ldots, N_e\}$ is obtained by calculating the unknown disturbance vector $w$ that will minimize the deviations between the estimates and the actual measurements. Towards that, a nonlinear optimization problem is formulated for the estimation window with the following cost function:

$$J_{est} = \sum_{i=0}^{N_e} \underbrace{(\hat{x}_{k-i|k} - y_{k-i})^T Q_e (\hat{x}_{k-i|k} - y_{k-i})}_{\text{estimation deviation}} \\ + \underbrace{(w_{k-i|k} - w_{\text{pr}})^T Q_{ch} (w_{k-i|k} - w_{\text{pr}})}_{\text{disturbance rate cost}} \\ + \underbrace{w_{k-i|k}^T R_e w_{k-i|k}}_{\text{state disturbance cost}} \quad (22)$$

where $Q_e, R_e, Q_{ch} \in \mathbb{R}^8$ are gain matrices, $w_{\text{pr}} \in \mathbb{R}^8$ is the estimation of the disturbance in the previous execution of the estimation process and the subscript $\cdot_{k-i|k}$ denotes the model value of $\cdot$ for the time-step $k-i$ computed at time-step $k$. The estimation deviation term aims to minimize the deviation between the estimates and the actual measurements throughout the whole window so that the updates model is as close as possible to the reality, the disturbance rate cost term is used to limit sudden changes of the disturbance estimates that may result in undesired oscillatory behavior and the state disturbance cost is used to minimize the overall disturbance estimate.

Overall, the optimization problem used in the disturbance estimation is process is shown below:

$$\begin{aligned}
\arg\min_{w} \quad & J_{est}(\hat{X}, Y, w, w_{\text{pr}}) \\
\text{s.t.} \quad & x_{k-i|k} = f(x_{k-i-1|k}, u_{k-i-1|k}) + w_{k-i|k} \\
& w_{k-i|k} = w_{k-i-1|k} \\
& w_{min} \leq w_{k-i|k} \leq w_{max}
\end{aligned} \quad (23)$$

where $w_{min}, w_{max} \in \mathbb{R}^8$ correspond to lower and upper bounds for the state disturbance. The aforementioned non-linear minimization problem is solved onboard using OpEn [36] in similar fashion to the solution for the MPC mentioned in Section 8.3.

Finally, in each iteration of the control loop in time-step $k$, the model used for the prediction step of the MPC of Eq. 16 is updated with the estimated disturbance $w_{k|k}$ and the next control input $u_{k+1|k}$ is calculated based on it. These dynamic updates of the model in real time are enabling it to adapt to changes of the environment but also to the gradual weight loss due to the continuous material extrusion while flying.

## 9. Hardware Setup

In implementing the proposed framework, apart from the aerial vehicle itself, two key factors significantly influence the construction parameters, results, and overall quality. One is the selection of the material carried by the UAV, as its deposition in layers determines the formation of the given mesh. The material's properties, such as density and deposition behavior, play a crucial role in shaping the design constraints for both the extruder and the aerial robotic platform. At the same time, the limitations of the UAV, including its hardware capabilities and payload capacity, impose restrictions on material selection. This creates a trade-off, where the choice of material must balance its desired characteristics with the practical constraints of the aerial platform.

*9.1. Material*

The material must fulfill two essential criteria, being lightweight for compatibility with the constrained payload of flying robots, and possessing an optimal volume-to-mass ratio to maximize coverage for a specified carried mass. Thus, aligning with previous iterations [11], [12] an expandable foam is selected to be the main construction material of the proposed framework. Specifically, a commercial[1] single compound polyurethane foam is utilized and is contained in a pressurized commercial canister. Upon extrusion, it promptly engages with the surrounding air, resulting in a foamy texture. Concurrently, it initiates an expansion process until achieving full hardness and curing. It must be noted that the original material selected for the process was an off-the-shelf dual compound foam as well, with greater expansion features. However, it was discarded due to safety concerns, given the fact that it contained the chemical substance diisocynate in quantities exceeding the safety standards for indoor use.

*9.2. Extruder*

In order to handle the deposition action of the material, a modular 3D-printed mechanism is designed and assembled that is encapsulated on the top and bottom side of the cannister as shown in Fig. 10. The extruder cannister along with its mounting on the drone and valve to control it are depicted in Fig. 11 Specifically, the canister's top is linked directly to a brass ball valve to regulate the flow of the deposited material, propelled by the contained propulsion gases. The valve's control leaver is mechanically linked, to the servos output shaft, which governs the material flow. A 3D-printed bracket affixed to the canister's bottom secures the servo in place. The servo's rotational movement is controlled directly by the onboard computer, enabling real-time activation and deactivation of material extrusion, seamlessly integrating it into the proposed framework. At the tip of the extruder lies a small plastic pipe, affixed to the opposite end of the valve, completing the mechanism for material deposition.

---

[1]Sika Boom®-461 Top [40]

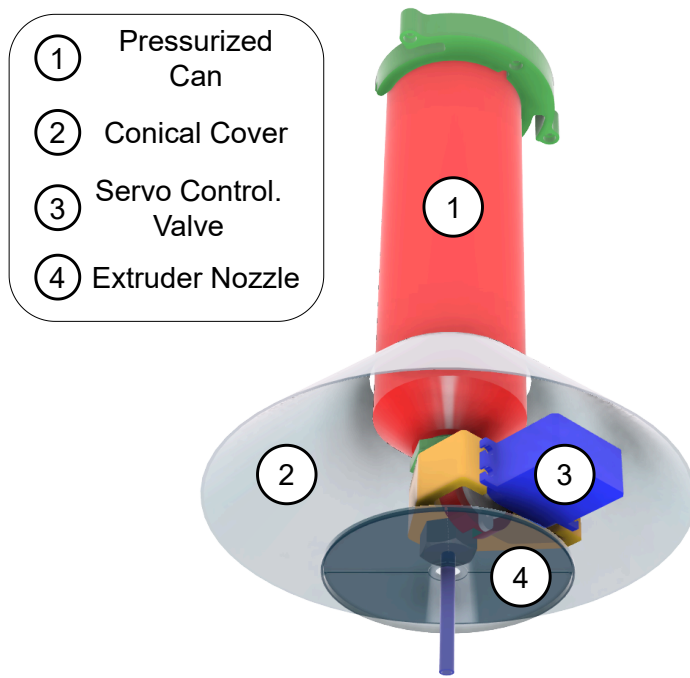

Figure 10: CAD Design of the custom material extruder and cannister holding it, affixed on the bottom part of the UAV.

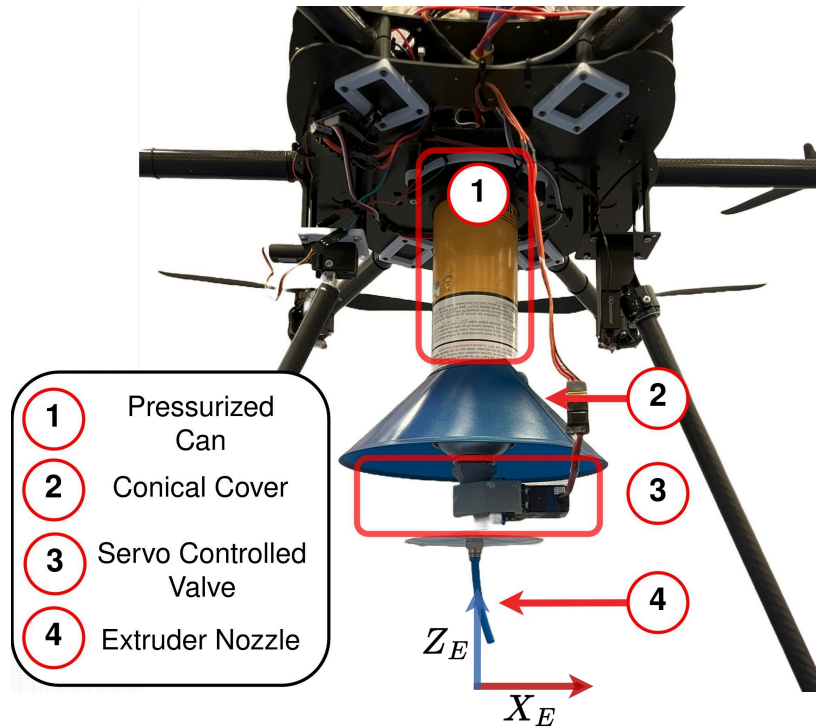

Figure 11: UAV platform designed with the extruder mounted below it equipped with a servo-controlled valve and protective disks reducing the downwash effect.

*9.3. Robotic Platform*

The whole extruder mechanism is mounted on the bottom side of an in-housed designed and assembled hexacopter aerial robotic platform with a wingspan of 0.8 meters, mass of 4.5 kilograms (batteries included) and designed to accommodate a payload of up to 2 kilograms with a flight time of approximately 10 minutes under full load. The platform is also equipped with retractable landing gear legs that are retracted when airborne due to the fact that the extruder needs to be as close to the printing surface as possible and also to prevent any possible collisions with the previously manufactured chunks. Finally, aiming to mitigate the effect of the downwash coming from the rotation motion of the propeller and affecting dramatically the material deposition behavior, a set of protective cover and disk is mounted right before and after the valve so that the flow is blocked. The degradation due to the phenomenon is not fully mitigated but is limited significantly.

## 10. Experimental Results

The proposed framework undergoes testing and evaluation within an indoor environment, where two distinct meshes were manufactured. Specifically, the experimentation took place within an indoor flying arena equipped with a Vicon motion-capturing system, providing the UAV with real-time odometry at a frequency of 120 Hz. Prior to commencing the manufacturing process, initial procedures such as chunking, slicing of generated chunks, and computation of the dependency graph were performed offline on a master PC. For the chunking algorithm beam search, an inner beam width $W_{inner} = 10$ and an outer one $W_{outter} = 10$ were selected while at the same time, each planar family for a given normal $\vec{n}_i$ consisted of 5 planes and the maximum polar angle of the normal sampling space was set to $\phi_{\max} = 45°$. After conducting handheld extrusion tests and experiments, the tuning parameters for the slicing software were determined. Specifically, a concentric infill pattern with 100% infill was chosen for filling the layers of the chunks. This approach minimizes the number of points where movement direction changes, leading to smoother and more efficient extrusion.

Subsequently, the remaining execution, movement, and control modules of the framework were executed on the onboard computational unit of the UAV, which utilized a LattePanda 3 Delta[2]. Additionally, the onboard Model Predictive Controller (MPC) was operated at a frequency of 20 Hz while the parameters of the model, the constraint bounds and the gains of its cost function are shown in Table 1.

The onboard Non-Linear Moving Horizon Estimator (NMHE) was executed in the same rate as the MPC and in each time step the model of the MPC was updated with the disturbance estimate $w$ of the estimator. The parameters used for the formulation of the NMHE problem are shown in Table 2.

### 10.1. Offset-Free MPC

To evaluate the offset-free MPC scheme described in Sections 8.3 and 8.4, a predefined path (shown in blue in Fig. 12) is tracked. Initially, this path is followed without the disturbance estimation module enabled, with the MPC using only the nominal model for predictions (see Fig. 13a). Although the overall pattern and corners of the path were tracked, offsets of 8cm on the y-axis and 3 cm on the x-axis are observed.

---

[2]https://www.lattepanda.com/lattepanda-3-delta

| Param. | Value | Param. | Value |
|---|---|---|---|
| A | [0.1, 0.1, 0.2] | $Q_u$ | $\text{diag}(3, 250, 250)$ |
| $\tau_\varphi$ | 0.2 | $Q_{\Delta u}$ | $\text{diag}(3, 100, 100)$ |
| $\tau_\theta$ | 0.2 | $u_{min}$ | [3, -0.3, -0.3] |
| $\delta_t$ | 50 ms | $u_{max}$ | [15.5, 0.3, 0.3] |
| N | 40 | $\Delta\varphi_{max}$ | 0.04 |
| $Q_x$ | $\text{diag}(100, 100, 50, 4, 4, 4, 30, 30)$ | $\Delta\theta_{max}$ | 0.04 |

Table 1: NMPC Controller Parameters

| Parameter | Value |
|---|---|
| $Q_e$ | [ $10e^4$, $10e^4$, $10e^4$, $10e^4$, $10e^4$, $10e^4$, $10e^4$, $10e^4$ ] |
| $Q_{ch}$ | [ 10, 10, 10, 10 ] [ 10, 10, 10, 10 ] |
| $R_e$ | [ 500, 500, 500, 1 ] [ 1, 1, 200, 200 ] |
| $W_{min}$ | [ $-0.05, -0.05, -0.05, -0.6$ ] [ $-0.6, -0.6, -0.05, -0.05$ ] |
| $W_{max}$ | [ 0.05, 0.05, 0.05, 0.6 ] [ 0.6, 0.6, 0.05, 0.05 ] |

Table 2: Parameter values for the Non-Linear Moving Horizon Estimator (NMHE).

At a specific timestamp $t_e$, the disturbance estimator is enabled mid-flight in order to showcase the enhancement in the tracking performance of the UAV. A disturbance of approximately 0.15 $m/s^2$ on the x-axis and 0.3 $m/s^2$ on the y-axis is estimated. The deviation from the predefined path is gradually decreased to within 2 cm (as shown in Fig. 13b), as the MPC model dynamically adapts to real flight conditions. The estimated disturbances on the x and y axes are shown in the figure. It is noted that only the disturbances on the acceleration part of the model are used, estimating virtual forces on each axis. The path measured before the estimator is deployed is marked in orange, and the path after deployment is marked in green in Fig. 12.

Although variations in the disturbance estimates occurred, as expected due to the dynamic nature of the estimation process, they do not significantly affect the overall mission performance due to the expansive nature of the foam.

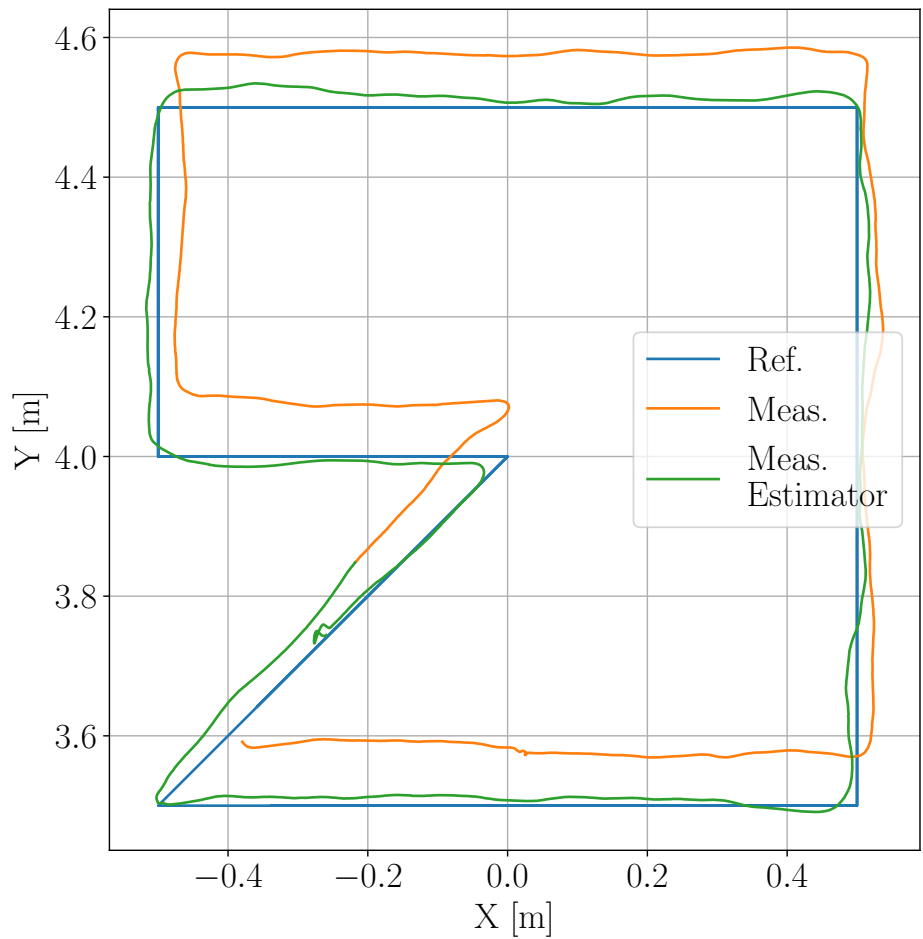

Figure 12: Reference Path for a single chunk (blue) along with the measured trajectory of the UAV before (orange) and after (green) the deployment of the estimator during flight.

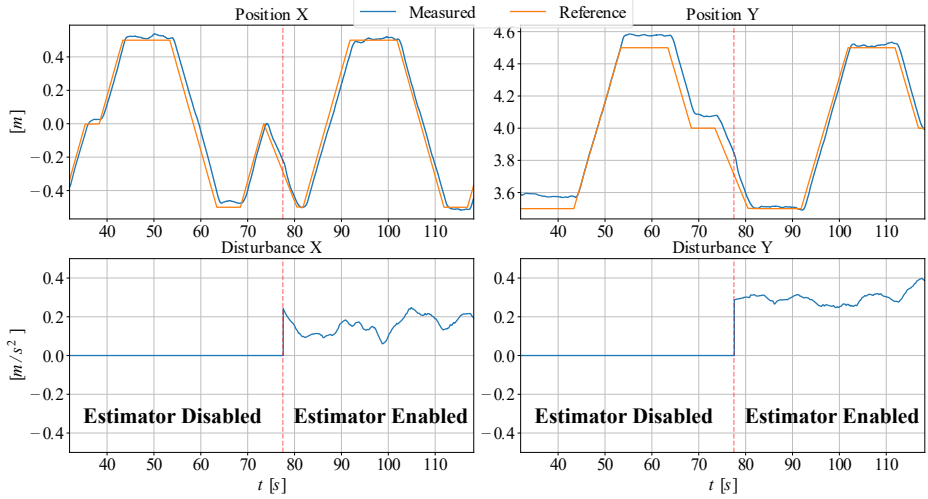

(a) A snapshot of reference and measured position in $X-Y$ coordinates (top) on the execution path along with estimated disturbances (bottom)

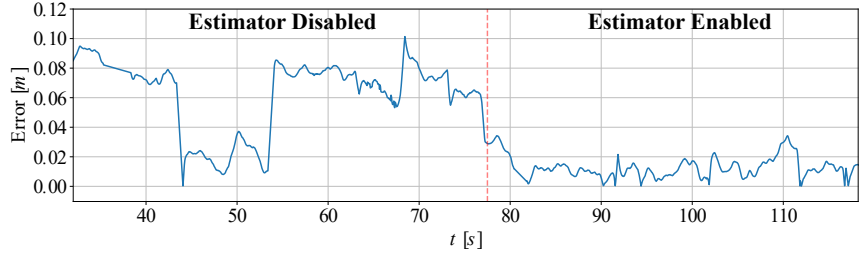

(b) Planar path tracking error, showing reduced error with the disturbance estimator enabled (right) versus disabled (left).

Figure 13: Snapshot of path tracking where disturbance estimator is enabled mid-flight ($t_e = 77$ sec) and the timestamp is indicated with the red dotted line. The response of the UAV along with the reference paths in each axis along with the estimated disturbances per axis are shown on top (a). The 2D error of the UAV during the tracking is shown on the bottom and is consistently decreased after the deployment of the estimator (b).

## 10.2. Construction of a Rectangular Shape

The framework is initially evaluated on the manufacturing of a solid $55 \times 55$ cm. rectangle with a height of 8 cm, having a hole in its center with dimensions of $20 \times 20$ cm. The available material for the UAVs that was fed to the chunking algorithm is as follows $\mathcal{D} = [4.0, 4.0, 4.0, 4.0]$ L while the layer height along with the line width used during the slicing process are equal to $l_h = 2.5$ cm and $l_w = 3$ cm. After feeding it to the chunking algorithm, the mesh is decomposed in 8 chunks as shown in Fig. 14 with the following volumes $\mathcal{C} = [1.90, 1.09, 2.51, 2.01, 2.57, 2.06, 2.35, 2.31]$ L with a mean value $\mu = 2.1$, standard deviation $\sigma = 0.442$ and a total volume dispersion $c_v = 0.211$.

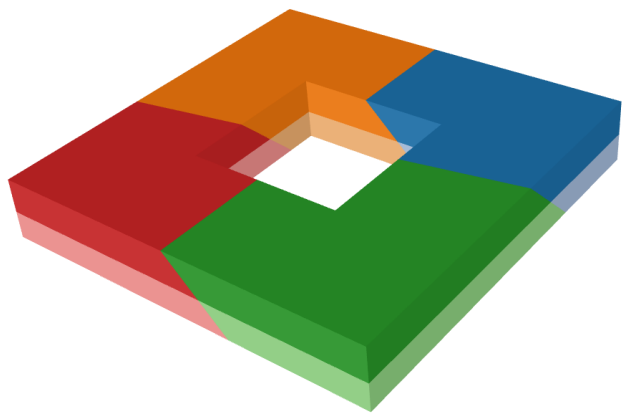

(a) Color-Coded Decomposed Mesh

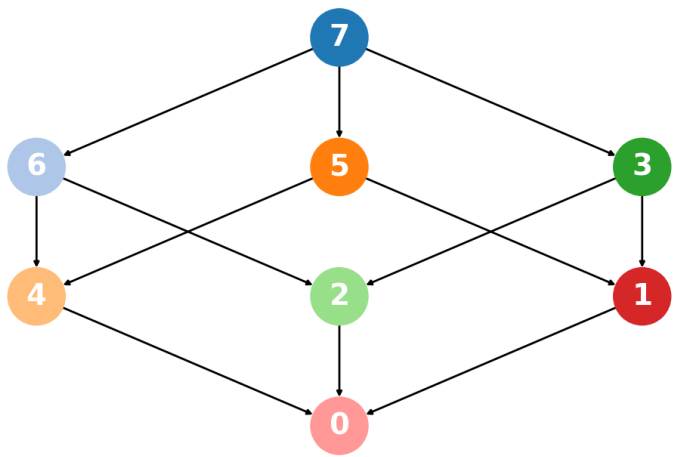

(b) Dependency Graph $\mathbf{G}$

Figure 14: Color-coded chunks of the decomposed mesh to be printed (a) along with the dependency graph showcasing all the inter-dependencies between the chunks (b).

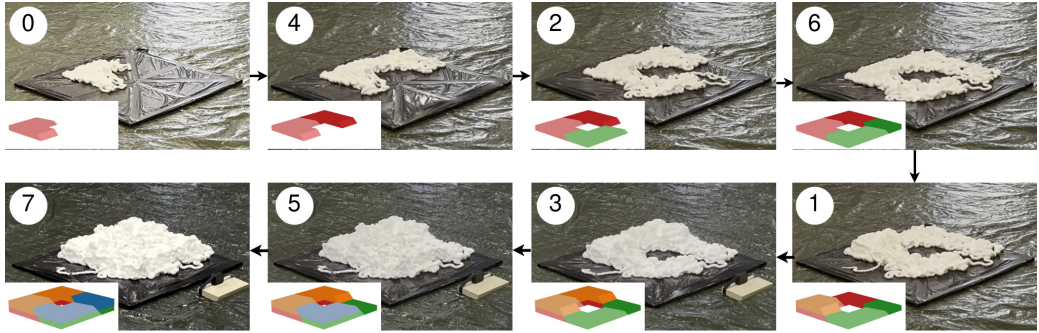

Figure 15: Sequential Manufacturing of the Chunks of the Rectangular Decomposed Mesh. The chunk that is printed in each frame is notated with the corresponding color and its number is highlighted on the top left.

Sequential snapshots of the overall manufacturing process taken right after the completion of each assigned chunk to the UAV are shown in Fig. 15. The final constructed mesh after having the material fully cured is shown in Fig. 16 in different views. As noticed from the figure, while the material extruded from the tip of the extruder of the robot is deposited according to the manufacturing paths and there is sufficient adhesion both between material and the ground platform and the intermediate layers, the deposited material is following a serpentine-like pattern while deposited. This can be accounted to the fact that material is not released with enough pressure so it starts expanding midair before it reaches either the ground or the previously deposited layer. This phenomenon can be controlled further by adjusting the valve opening through the attached servo on it.

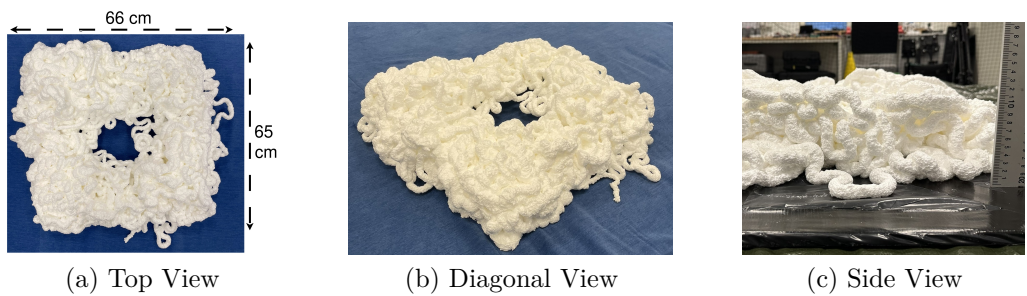

(a) Top View  (b) Diagonal View  (c) Side View

Figure 16: Final Manufactured Rectangular Mesh after fully curing

*10.3. Construction of a Hexagonal Shape*

Additionally, the manufacture of a hexagonal mesh is carried out and evaluated. Each side of it has length of 30 cm and width of 10 cm while the height is equal to 10 cm resulting in a total volume of 18.93 L. Four different material canisters were considered available during the mesh decomposition process with 6 L of material each so the vector $\mathcal{D}$ is as follows $\mathcal{D} = [6.0, 6.0, 6.0, 6.0]$ L.

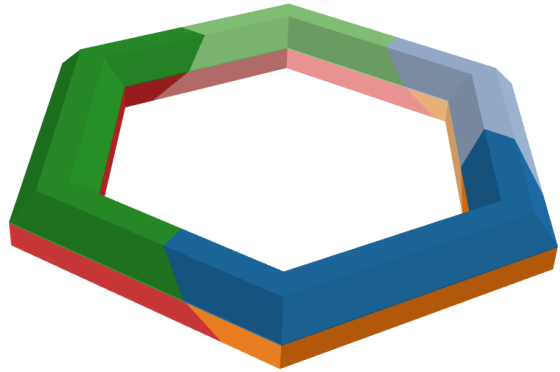

(a) Color-Coded Decomposed Mesh

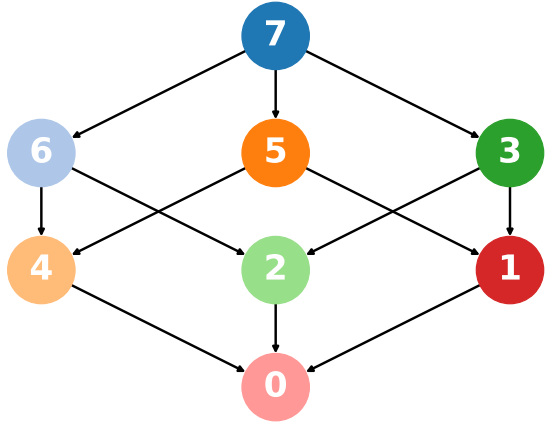

(b) Dependency Graph **G**

Figure 17: Color-coded chunks of the decomposed hexagonal mesh to be printed (a) along with the dependency graph showcasing all the inter-dependencies between the chunks (b).

A layer height $l_h = 4.5$ cm was used and a line width of 4 cm as well.

The mesh was decomposed into 8 chunks with the following volumes $\mathcal{C} = [2.39, 1.34, 2.57, 1.79, 2.77, 1.71, 3.58, 2.79]$ L corresponding to each chunk $C_k$, resulting in mean value $\mu = 2.3675$, standard deviation $\sigma = 0.678$ and a total volume dispersion $c_v = 0.306$. The chunks of the decomposed hexagonal mesh are shown color-coded in Fig. 17a while the dependency graph associated with it is shown in Fig. 17b. An overview of the sequential assignment of the chunks based on the assignment algorithm analyzed in Section 6.1 is shown in Fig. 18, where multiple snapshots of the printing mission are presented in a chronological order and each frame is captured right after the UAV has finished printing the chunk corresponding to it. It is noted that although the chunks can be printed in various sequences, the selection of the chunks is happening in layers $\mathcal{L}_\mathcal{G}^1 = \{C_0, C_2, C_4, C_6\}$ and $\mathcal{L}_\mathcal{G}^2 = \{C_1, C_3, C_5, C_7\}$ as dictated in the aforementioned Section. Another snapshot view for the printing of a single chunk is shown in Fig. 19, where a camera is placed next to the printing platform and the deposition action of the material from the extruder is captured. The final constructed hexagonal mesh after letting the material fully cure along with the dimension of the original one are shown in Fig. 20. As shown, the overall constructed shape closely resembles the original structure, with dimensions that are largely maintained. However, some regions exhibit uneven material expansion after depletion, resulting in small gaps. These gaps are caused by the influence of the downwash generated by the propellers, which is further discussed in Section 11. It is important to note that this experimental evaluation aims to demonstrate the proof of concept. To this end, the targeted shapes for printing are relatively small, making execution highly challenging. However, a similar margin of error can be expected even for larger targeted shapes leading to much less noticeable artifacts, ensuring smoother and consistent results as the system scales

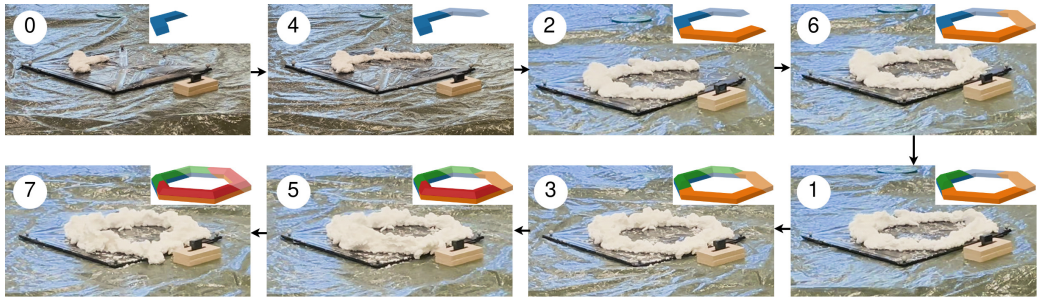

Figure 18: Sequential Manufacturing of the Chunks of the Hexagonal Decomposed Mesh. The chunk that is printed in each frame is added on the visualization on the top right of each frame and its number is highlighted on the top left.

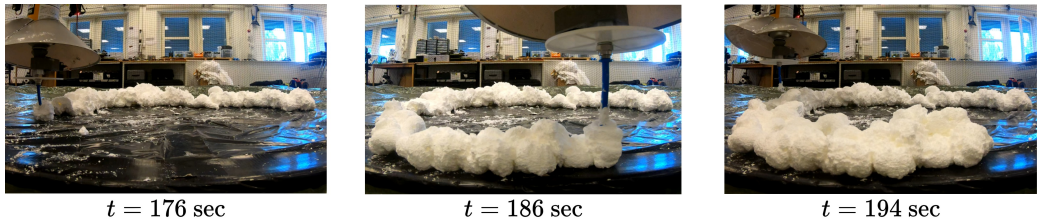

$t = 176$ sec $\qquad$ $t = 186$ sec $\qquad$ $t = 194$ sec

Figure 19: Side view of snapshots taken during the printing of the second chunk in the printing sequence for the hexagonal.

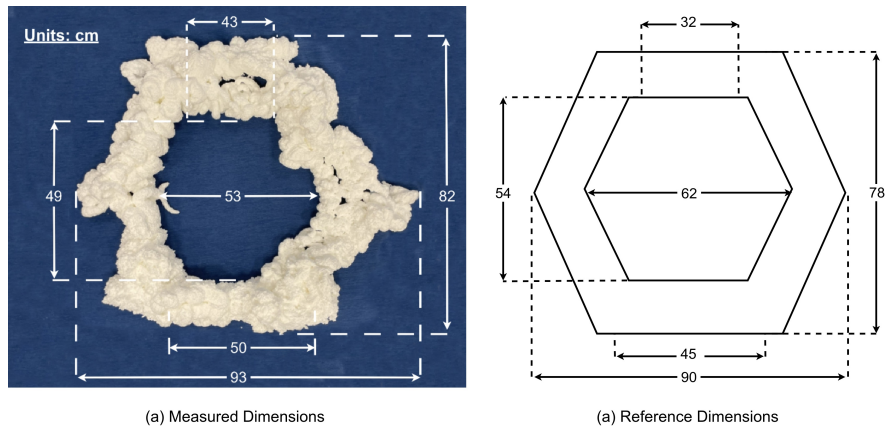

(a) Measured Dimensions $\qquad$ (a) Reference Dimensions

Figure 20: Dimensions and various length sizes annotated for both the finally constructed (a) and the reference (b) Hexagonal Mesh.

A presentation of the whole framework along with videos from the two aforementioned experiments can be found here: https://youtu.be/WC1rLM-LKEg4.

## 11. Discussions

A thorough discussion of the experimental results follows focusing on the accuracy of the whole printing procedure and evaluation on the accurate representation of the digital fabrication based on volumetric occupancy grids. Additionally, a discussion on phenomena resulting to the degrading of the printing mission and ways of mitigating them.

*11.1. Printing Execution*

The tracking of the measured position of the UAV as fed from the motion capture tracking system along with the reference path are shown in Fig. 21 for the printing of various chunks separated with black dotted line. Given the fact that the extruder of the platform is rigidly mounted at the bottom of the canister with a vertical offset equal to $l_{ex} = 30$ cm and the pose of the robot is known precisely in real-time, a calculation of the extruder is carried out so that the tip of it is measured at every measurement while the measured and the reference positions of the extruder are shown at the same figure. As noticed from the response in each axis, the paths are tracked with a slight lag between the reference signal and the response which in the particular case of the initial manufacturing tests is considered acceptable given the fact that it does not affect the accuracy or the smooth deposition of the material. Slight deviations in the UAV's commanded reference altitude can be observed in the lower portion of the figure, with variations within a range of 2.5 cm. Although this behavior does not affect the material deposition process, it is accounted for in the UAV's execution. Specifically, the extruder's tip is positioned with an offset that ensures it remains clear of both the ground and the previously deposited material. This prevents ground contact, even under worst-case deviations, as the material is expelled under pressure from the canister, allowing for effective deposition without requiring precise tip-to-ground contact, similar to a spraying action. The trajectory of the extruder exhibits slight fluctuations around the reference signal, primarily due to its vertical position relative to the UAV's center of mass. This offset means that even small angular deviations of the UAV can introduce amplified movements at the extruder tip. By compensating for these rotational influences, the system can achieve smoother extrusion paths, ultimately enhancing precision. This challenge also highlights the importance of robust stabilization techniques, which, when optimized, further improves printing accuracy.

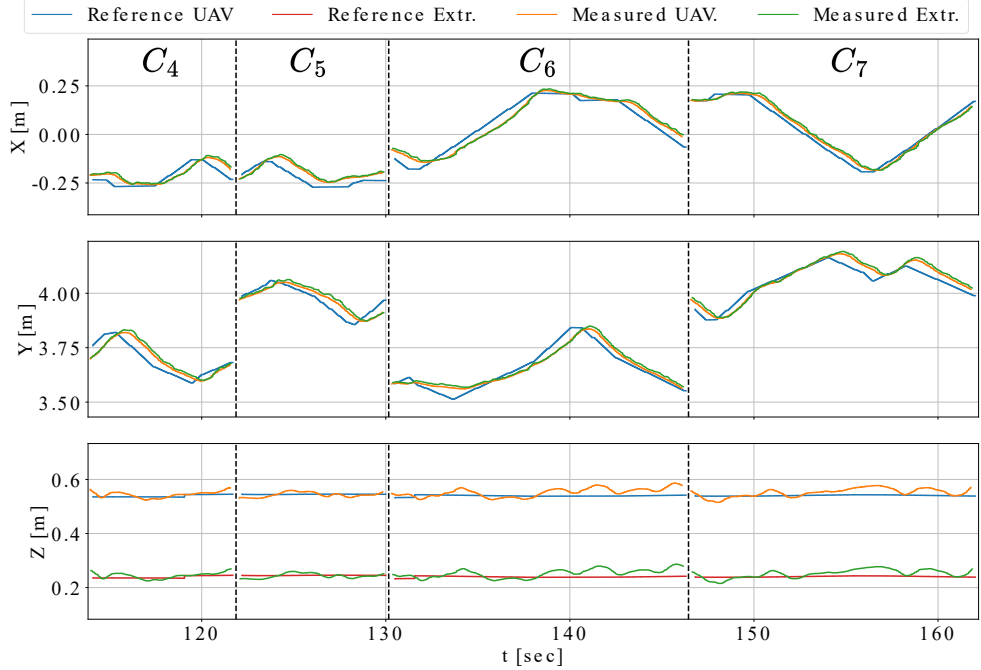

Figure 21: Reference and measured trajectory of both UAV an extruder per axis for different chunks separated in time using the black dotted line

Another factor affecting the accuracy of the manufacturing path tracking is the ground effect. During material extrusion, the UAV flies at a vertical distance of approximately 30 cm from the ground, which is a critical altitude for the system to be influenced by this phenomenon [41]. The disturbance estimator effectively accounts for ground effect by calculating overall unmeasured disturbances due to its reactive nature. This ensures reliable compensation, leading to a robust control system that supports high precision and consistency in tracking the intended path, thereby enhancing printing accuracy. More detail insights and analysis on robustness properties of such controller can be found in [42, 43].

A better 3D view on the actually tracked trajectories for both hexagonal and rectangular meshes is shown in Fig. 22. Here, the vertical offset between the UAV's and the extruder's path is more visible and the motion of the extruder relative to the UAV is shown clear. A holistic overview of the total error for both UAV and extruder's tip is shown in Fig.23 where the distribution of the mean error is visualized for the printing of each chunk throughout a whole mission. Notably, the mean error of the UAV is around 2.1 cm while it never reaches over 5 cm. As expected, the distribution of the error of the extruder is slightly worse with a mean error of 2.5 cm and an upper bound of 6 cm.

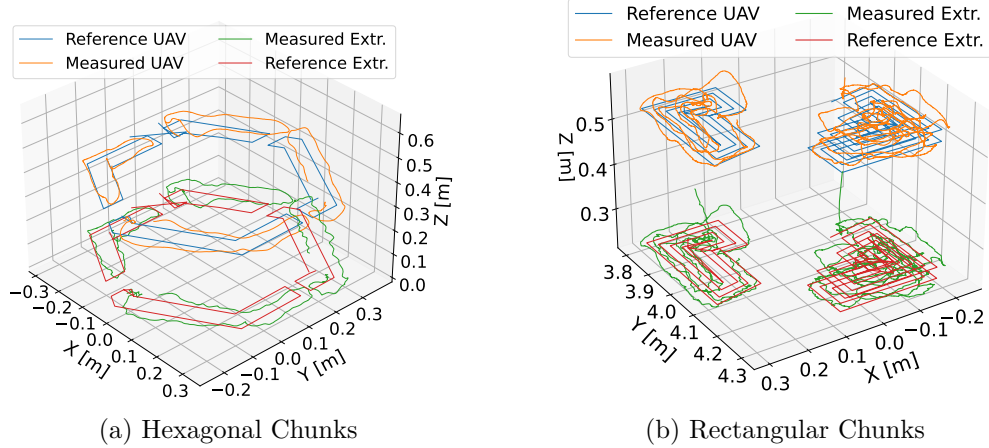

(a) Hexagonal Chunks  (b) Rectangular Chunks

Figure 22: Reference and measured trajectories of both UAV and extruder's tip in 3D during the printing process of the chunks of the hexagonal mesh(a) and of the chunks of the rectangular mesh(b).

It must be noted that in larger-scale constructions, where the relative error diminishes, these deviations become even less significant, further reinforcing the reliability of the approach.

*11.2. Volumetric Evaluation*

On further evaluation of the printing execution of the UAV, a volumetric representation of the manufacturing paths generated by the slicer and the actual trajectory executed by the UAV are compared. Specifically, a 3D occupancy grid is generated with dimensions equal to the ones of the printing platform and a height of 0.5 m. Aiming to capture accurately the aforementioned trajectories, a voxel size equal to 1 cm is used. This step is carried out for both the measured and the reference paths resulting into two occupancy maps $O_r$ and $O_m$ correspondingly. As shown in Fig. 24, the manufacturing paths are visualized with green and whenever the extrusion is enabled, the measured position of the extruder is sampled, padded to dimensions equal to the line width $l_w$ and registered in the occupancy grid.

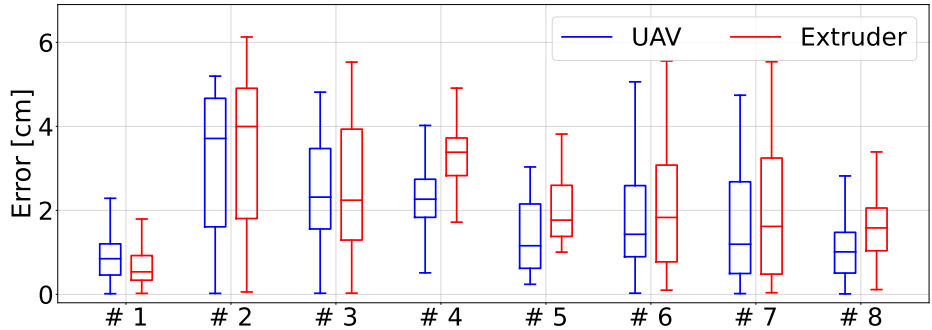

Figure 23: Path tracking error while printing all chunks for both the UAV and the extruder.

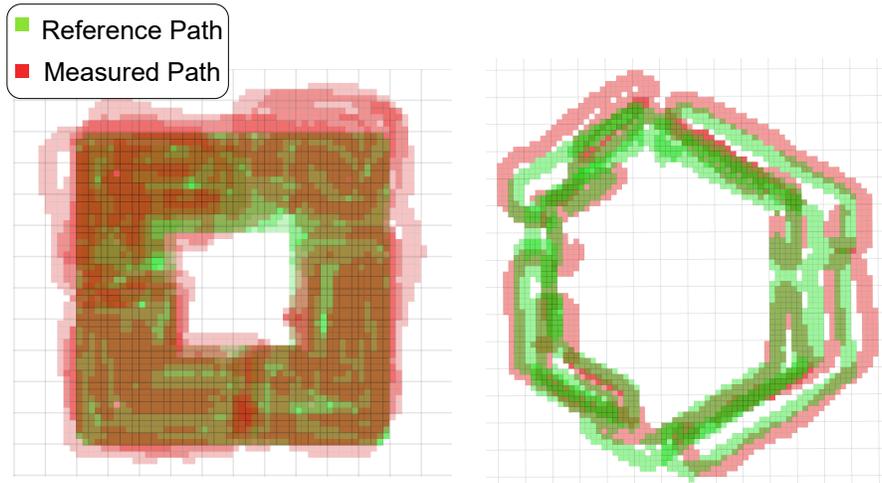

Figure 24: Volumetric evaluation of the manufactured meshes in an occupancy grid of voxel size equal to 1 cm

*11.3. Air Draft Influence*

As it can be noticed on both the figures shown so far along with the accompanying video, the lightweight nature of the foam used in the printing process, along with its extrusion in mid-air, makes the final deposition position sometimes susceptible to air drafts. Specifically, the rotating propellers

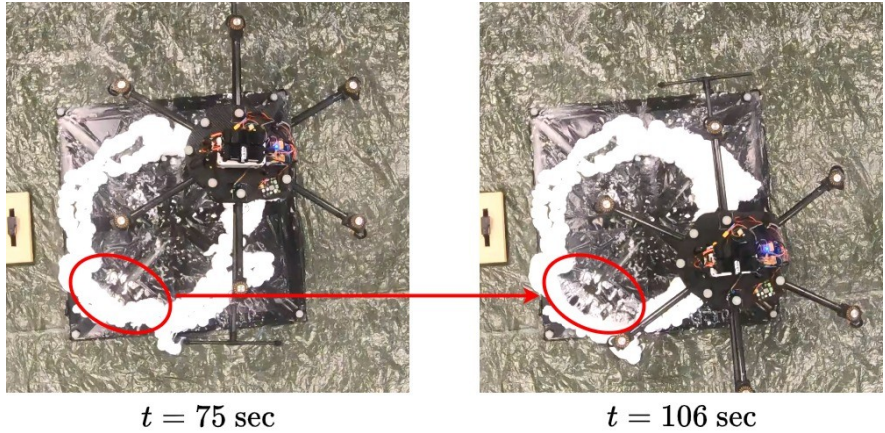

$t = 75$ sec $\qquad\qquad t = 106$ sec

Figure 25: Influence of aircraft on chunk $C_4$ (marked with red) generated by the propellers when printing a nearby chunk.

of the UAVs generate downward airflows, which, when combined with the proximity of the UAV to the ground, create unpredictable air patterns due to reflections off the surface. These air vortices interfere with the material deposition, often displacing the foam before it cures, making successful deposition impossible as the material is blown away. To address this, a conical cover and a protective disk were added above the extruder's tip, as shown in Fig. 11. This modification significantly reduced the influence of the drafts, allowing the printing process to proceed.

Nevertheless, air drafts generated by the UAVs' propellers continue to occasionally affect the deposition performance and the quality of the final print. This issue becomes particularly apparent when the UAV operates near a recently printed chunk, where the newly deposited layers have not yet formed a strong adhesive bond with the substrate or previously deposited sections. Throughout the experimentation of the printing of the hexagonal mesh, this phenomenon is particularly prominent after the fabrication of chunk $C_4$, the final chunk in the first layer of the hexagonal mesh. Once this chunk is printed, the structure becomes enclosed, preventing the air drafts from escaping. As a result, these trapped drafts exert pressure on the internal walls of the surrounding chunks. as also illustrated in Fig. 25. Although the displacement of the chunk material is noticeable, it does not critically compromise the structural integrity of the mesh, as the remaining material is sufficient to support the subsequent layers of printing.

*11.4. Material Extrusion*

Throughout the mission, the extrusion process was successfully controlled by adjusting the valve's angle using a servo motor, effectively initiating material flow. This angle was carefully determined through preliminary handheld experiments, providing a solid foundation for the autonomous mission. While some variations in extrusion were observed, they offered valuable insights into optimizing flow control. Specifically, at the start of extrusion, a temporary pressure buildup led to a higher initial flow rate before stabilizing. Similarly, as the mission progressed, a gradual decrease in pressure within the canister caused a corresponding reduction in flow rate. Understanding these dynamics presents an exciting opportunity for refinement, paving the way for even more precise and uniform deposition. To further enhance consistency, integrating a flow sensor into the valve could enable a feedback control system that dynamically adjusts the servo, ensuring a steady flow rate.

## 12. Conclusions and Future Work

This article presents a chuck-based aerial additive manufacturing framework, supported by experimental evaluation, marking a significant advancement in enhancing aerial robotic capabilities for autonomous construction. The experimental evaluation pushes the boundaries of this framework, driving it toward a transformative shift while addressing the critical challenges of deploying aerial robotic units for construction tasks. This study not only highlights the intricate execution challenges of deploying aerial robotic systems for construction but also presents the solutions and insights gained from our experimental experience. Towards this development, the chuck based framework is augmented with a dependency graph structure that captures all inter-dependencies between the chunks. This facilitates the efficient assignment of each chunk to the available UAVs. Additionally, a more refined chunking process reduces excess artifacts, while a post-processing algorithm in chunk generation improves adhesion between chunks and ensures their full inclusion in the slicing process. A real-time disturbance estimation module integrated with the controller ensures smooth and precise execution of the printing process. An in-house designed and assembled hexacopter equipped with a pressurized canister mounted on its bottom part is deployed as the aerial robotic construction unit used in the experiments. Using this platform, the experimental evaluation demonstrates the aerial robotic capability in autonomously constructing two distinct mesh structures, one of them being a hollow rectangle and the other one a hexagonal mesh, highlighting its versatility and effectiveness in aerial additive manufacturing. During the execution of the construction process, the material carrying canister was replaced for every landing sequences so that a sequential manufacturing of the chunks are carried out seamlessly. A rigorous evaluation of the constructed meshes was conducted, focusing on geometric properties such as shape size and volume. Limited discrepancies were observed, which is trivial in experimental validation and it can be attributed to factors such as uncompensated disturbance noise, model inaccuracies, and the ground effect due to the close proximity of the flying platform to the ground. Additionally, in spite of the favorable properties of the material being lightweight and being able to expand, its extrusion behavior was not consistent and smooth throughout the whole execution of the printing mission since variations in its flow rate, extrusion pattern and expansion degraded the final printing quality. While the constructed shapes using aerial robotic system shows marginal errors, it is majorly due to the fact that the experimental evaluation are performed towards constructing small structures. Considering that the performance of the tracking controller is expected to be independent of the execution path length, the effect of small-scale tracking errors would be insignificantly pronounced when deployed on a larger scale. When constructing larger structures, the relative error margin would remain bounded within acceptable tolerance in the overall construction process. It is evident that the proposed approach successfully validated the proof of concept demonstration.

To enhance printing performance in future development, opting for a more suitable cement like construction material with superior properties could address previously mentioned challenges and improve overall mission efficiency. Furthermore, deploying multiple UAVs to manufacture their assigned chunks simultaneously, while managing conflicts arising from concurrent motion, will significantly accelerate the overall process. Additionally, an elaborate computational fluid dynamics (CFD) analysis on the extrusion of the material from the canister along with its deposition in layers, could be further investigated and provide further insights improving the overall printing quality. Finally, augmenting aerial 3D printing capabilities with conventional well-developed multi-robotic system for a full-scale, real-life implementation of multi-agent and multi-modal collaborative construction would significantly enhance the envisioned evolution of autonomy in construction domain.